\documentclass[letterpaper, 10 pt, conference]{ieeeconf}  %
\usepackage{times}

\usepackage{multicol}
\usepackage[bookmarks=true]{hyperref}
\usepackage{xcolor}
\usepackage{color}
\usepackage{graphicx}
\usepackage{siunitx}

\usepackage{amsmath}
\usepackage{amssymb}
\usepackage{url}                 
\usepackage{graphicx}
\usepackage{graphics}
\usepackage{pifont}
\usepackage{color}
\usepackage{multirow}
\usepackage{adjustbox}
\usepackage{subcaption}
\usepackage{bm}
\usepackage{algorithm} 
\usepackage{algpseudocode}
\usepackage{wrapfig}
\usepackage{fancybox}
\usepackage{tikz}
\usepackage{subcaption}
\usepackage{microtype}      %
\usepackage{booktabs,subcaption,amsfonts,dcolumn}
\usepackage{graphicx}
\usepackage{adjustbox}
\usepackage{tabularx}
\usepackage{diagbox}
\usepackage{makecell}
\usepackage[font=footnotesize,labelformat=empty]{subcaption}
\usepackage{tikz}

\IEEEoverridecommandlockouts %
\title{\LARGE \bf
Sim-on-Wheels: Physical World in the Loop Simulation for Self-Driving
}
\author{Yuan Shen$^*$, Bhargav Chandaka$^*$, Zhi-hao Lin, Albert Zhai, Hang Cui, David Forsyth and Shenlong Wang%
    \thanks{* denotes equal contributions. The authors are with the University of Illinois at Urbana-Champaign.  
            Email: {\tt \{yshen47, bhargav9, cl121, azhai2, hangcui3, daf, shenlong\}@illinois.edu})
    }
}
\begin{document}

\newcommand{\todocite}[1]{\textcolor{blue}{Citation needed []}}
\newcommand{\shenlongsay}[1]{\textcolor{blue}{[{\it Shenlong: #1}]}}
\newcommand{\shenlong}[1]{{\color{magenta} #1}}

\newcommand{\todo}[1]{\textcolor{red}{[\textit{TODO: #1}]}}

\newcommand{\mfigure}[2]{\includegraphics[width=#1\linewidth]{#2}}
\newcommand{\mpage}[2]
{
\begin{minipage}{#1\linewidth}\centering
#2
\end{minipage}
}

\newcommand{\xpar}[1]{\noindent\textbf{#1}\ \ }
\newcommand{\vpar}[1]{\vspace{3mm}\noindent\textbf{#1}\ \ }

\newcommand{\sect}[1]{Section~\ref{#1}}
\newcommand{\sects}[1]{Sections~\ref{#1}}
\newcommand{\eqn}[1]{Equation~\ref{#1}}
\newcommand{\eqns}[1]{Equations~\ref{#1}}
\newcommand{\fig}[1]{Figure~\ref{#1}}
\newcommand{\figs}[1]{Figures~\ref{#1}}
\newcommand{\tab}[1]{Table~\ref{#1}}
\newcommand{\tabs}[1]{Tables~\ref{#1}}

\newcommand{\ignorethis}[1]{}
\newcommand{\norm}[1]{\lVert#1\rVert}
\newcommand{\fcseven}{$\mbox{fc}_7$}

\renewcommand*{\thefootnote}{\fnsymbol{footnote}}

\def\naive{na\"{\i}ve\xspace}
\def\Naive{Na\"{\i}ve\xspace}

\makeatletter
\DeclareRobustCommand\onedot{\futurelet\@let@token\@onedot}
\def\@onedot{\ifx\@let@token.\else.\null\fi\xspace}

\def\iid{\emph{i.i.d}\onedot}
\def\eg{\emph{e.g}\onedot} \def\Eg{\emph{E.g}\onedot}
\def\ie{\emph{i.e}\onedot} \def\Ie{\emph{I.e}\onedot}
\def\cf{\emph{c.f}\onedot} \def\Cf{\emph{C.f}\onedot}
\def\etc{\emph{etc}\onedot} \def\vs{\emph{vs}\onedot}
\def\wrt{w.r.t\onedot} \def\dof{d.o.f\onedot}
\def\etal{\emph{et al}\onedot}
\makeatother

\newcommand{\myparagraph}[1]{\vspace{-6pt}\paragraph{#1}}

\newcommand{\bbR}{{\mathbb{R}}}
\newcommand{\bK}{\mathbf{K}}
\newcommand{\bX}{\mathbf{X}}
\newcommand{\bY}{\mathbf{Y}}
\newcommand{\bk}{\mathbf{k}}
\newcommand{\bx}{\mathbf{x}}
\newcommand{\by}{\mathbf{y}}
\newcommand{\bhy}{\hat{\mathbf{y}}}
\newcommand{\bty}{\tilde{\mathbf{y}}}
\newcommand{\bG}{\mathbf{G}}
\newcommand{\bI}{\mathbf{I}}
\newcommand{\bg}{\mathbf{g}}
\newcommand{\bS}{\mathbf{S}}
\newcommand{\bs}{\mathbf{s}}
\newcommand{\bM}{\mathbf{M}}
\newcommand{\bw}{\mathbf{w}}
\newcommand{\eye}{\mathbf{I}}
\newcommand{\bU}{\mathbf{U}}
\newcommand{\bV}{\mathbf{V}}
\newcommand{\bW}{\mathbf{W}}
\newcommand{\bn}{\mathbf{n}}
\newcommand{\bv}{\mathbf{v}}
\newcommand{\bq}{\mathbf{q}}
\newcommand{\bR}{\mathbf{R}}
\newcommand{\bi}{\mathbf{i}}
\newcommand{\bj}{\mathbf{j}}
\newcommand{\bp}{\mathbf{p}}
\newcommand{\bt}{\mathbf{t}}
\newcommand{\bJ}{\mathbf{J}}
\newcommand{\bu}{\mathbf{u}}
\newcommand{\bB}{\mathbf{B}}
\newcommand{\bD}{\mathbf{D}}
\newcommand{\bz}{\mathbf{z}}
\newcommand{\bP}{\mathbf{P}}
\newcommand{\bC}{\mathbf{C}}
\newcommand{\bA}{\mathbf{A}}
\newcommand{\bZ}{\mathbf{Z}}
\newcommand{\bff}{\mathbf{f}}
\newcommand{\bF}{\mathbf{F}}
\newcommand{\bo}{\mathbf{o}}
\newcommand{\bc}{\mathbf{c}}
\newcommand{\bT}{\mathbf{T}}
\newcommand{\bQ}{\mathbf{Q}}
\newcommand{\bL}{\mathbf{L}}
\newcommand{\bl}{\mathbf{l}}
\newcommand{\ba}{\mathbf{a}}
\newcommand{\bE}{\mathbf{E}}
\newcommand{\bH}{\mathbf{H}}
\newcommand{\bd}{\mathbf{d}}
\newcommand{\br}{\mathbf{r}}
\newcommand{\bb}{\mathbf{b}}
\newcommand{\bh}{\mathbf{h}}

\newcommand{\btheta}{\bm{\theta}}
\newcommand{\bhh}{\hat{\mathbf{h}}}
\newcommand{\ci}{{\cal I}}
\newcommand{\ct}{{\cal T}}
\newcommand{\co}{{\cal O}}
\newcommand{\ck}{{\cal K}}
\newcommand{\cu}{{\cal U}}
\newcommand{\cS}{{\cal S}}
\newcommand{\cQ}{{\cal Q}}
\newcommand{\cT}{{\cal S}}
\newcommand{\cC}{{\cal C}}
\newcommand{\cE}{{\cal E}}
\newcommand{\cF}{{\cal F}}
\newcommand{\cL}{{\cal L}}
\newcommand{\X}{{\cal{X}}}
\newcommand{\Y}{{\cal Y}}
\newcommand{\cH}{{\cal H}}
\newcommand{\cP}{{\cal P}}
\newcommand{\cN}{{\cal N}}
\newcommand{\cU}{{\cal U}}
\newcommand{\cV}{{\cal V}}
\newcommand{\cX}{{\cal X}}
\newcommand{\cY}{{\cal Y}}
\newcommand{\graph}{{\cal H}}
\newcommand{\bayes}{{\cal B}}
\newcommand{\cx}{{\cal X}}
\newcommand{\cg}{{\cal G}}
\newcommand{\cM}{{\cal M}}
\newcommand{\cG}{{\cal G}}
\newcommand{\cR}{\cal{R}}
\newcommand{\R}{\cal{R}}
\newcommand{\eig}{\mathrm{eig}}

\newcommand{\bbS}{\mathbb{S}}

\newcommand{\D}{{\cal D}}
\newcommand{\bfp}{{\bf p}}
\newcommand{\bfd}{{\bf d}}

\newcommand{\cv}{{\cal V}}
\newcommand{\ce}{{\cal E}}
\newcommand{\cy}{{\cal Y}}
\newcommand{\cz}{{\cal Z}}
\newcommand{\cb}{{\cal B}}
\newcommand{\cq}{{\cal Q}}
\newcommand{\cd}{{\cal D}}
\newcommand{\bcf}{{\cal F}}
\newcommand{\cI}{\mathcal{I}}

\newcommand{\ut}{^{(t)}}
\newcommand{\up}{^{(t-1)}}

\newcommand{\bpi}{\boldsymbol{\pi}}
\newcommand{\bphi}{\boldsymbol{\phi}}
\newcommand{\bPhi}{\boldsymbol{\Phi}}
\newcommand{\bmu}{\boldsymbol{\mu}}
\newcommand{\bSigma}{\boldsymbol{\Sigma}}
\newcommand{\bGamma}{\boldsymbol{\Gamma}}
\newcommand{\bbeta}{\boldsymbol{\beta}}
\newcommand{\bomega}{\boldsymbol{\omega}}
\newcommand{\blambda}{\boldsymbol{\lambda}}
\newcommand{\bkappa}{\boldsymbol{\kappa}}
\newcommand{\btau}{\boldsymbol{\tau}}
\newcommand{\balpha}{\boldsymbol{\alpha}}
\def\bgamma{\boldsymbol\gamma}

\newcommand{\prox}{{\mathrm{prox}}}

\newcommand{\pardev}[2]{\frac{\partial #1}{\partial #2}}
\newcommand{\dev}[2]{\frac{d #1}{d #2}}
\newcommand{\dw}{\delta\bw}
\newcommand{\lab}{\mathcal{L}}
\newcommand{\unlab}{\mathcal{U}}
\newcommand{\ind}{1{\hskip -2.5 pt}\hbox{I}}
\newcommand{\ff}[2]{   \cf_{\prec (#1 \rightarrow #2)}}
\newcommand{\vv}[2]{   \cv_{\prec (#1 \rightarrow #2)}}
\newcommand{\dd}[2]{   \delta_{#1 \rightarrow #2}}
\newcommand{\ld}[2]{   \lambda_{#1 \rightarrow #2}}
\newcommand{\en}[2]{  \bD(#1|| #2)}
\newcommand{\ex}[3]{  \bE_{#1 \sim #2}\left[ #3\right]} 
\newcommand{\exd}[2]{  \bE_{#1 }\left[ #2\right]}

\newcommand{\se}[1]{\mathfrak{se}(#1)}
\newcommand{\SE}[1]{\mathbb{SE}(#1)}
\newcommand{\so}[1]{\mathfrak{so}(#1)}
\newcommand{\SO}[1]{\mathbb{SO}(#1)}

\newcommand{\poselow}{\xi}
\newcommand{\pose}{\bm{\poselow}}
\newcommand{\linpose}{\pose^\ell}
\newcommand{\cbpose}{\pose^c}
\newcommand{\rateparam}{v_i}
\newcommand{\bapose}{\bm{\poselow}_i}
\newcommand{\trackingpose}{\bm{\poselow}}
\newcommand{\rotlow}{\omega}
\newcommand{\rot}{\bm{\rotlow}}
\newcommand{\translow}{v}
\newcommand{\trans}{\bm{\translow}}
\newcommand{\hnorm}[1]{\left\lVert#1\right\rVert_{\gamma}}
\newcommand{\lnorm}[1]{\left\lVert#1\right\rVert}
\newcommand{\barate}{v_i}
\newcommand{\trackingrate}{v}
\newcommand{\imgpt}{\mathbf{u}_{i,k,j}}
\newcommand{\mappt}{\mathbf{X}_{j}}
\newcommand{\timet}[1]{\bar{t}_{#1}}
\newcommand{\mf}[1]{\text{MF}_{#1}}
\newcommand{\kmf}[1]{\text{KMF}_{#1}}
\newcommand{\Exp}{\text{Exp}}
\newcommand{\Log}{\text{Log}}

\newcommand{\shiftleft}[2]{\makebox[0pt][r]{\makebox[#1][l]{#2}}}
\newcommand{\shiftright}[2]{\makebox[#1][r]{\makebox[0pt][l]{#2}}}

\maketitle
\thispagestyle{empty}
\pagestyle{empty}
\begin{abstract}
We present Sim-on-Wheels, a safe, realistic, and vehicle-in-loop framework to test autonomous vehicles' performance in the real world under safety-critical scenarios. Sim-on-wheels runs on a self-driving vehicle operating in the physical world. It creates virtual traffic participants with risky behaviors and seamlessly inserts the virtual events into images perceived from the physical world in real-time. The manipulated images are fed into autonomy, allowing the self-driving vehicle to react to such virtual events. The full pipeline runs on the actual vehicle and interacts with the physical world, but the safety-critical events it sees are virtual. 
Sim-on-Wheels is safe, interactive, realistic, and easy
to use. The experiments demonstrate the potential of Sim-on-Wheels to facilitate the process of testing autonomous driving in challenging real-world scenes with high fidelity and low risk. Additional results and open-sourced code are available on our project page here: \url{https://sim-on-wheels.github.io/}.
\end{abstract}
\section{Introduction}
\begin{figure*}[t]
\centering
\includegraphics[width=\linewidth]{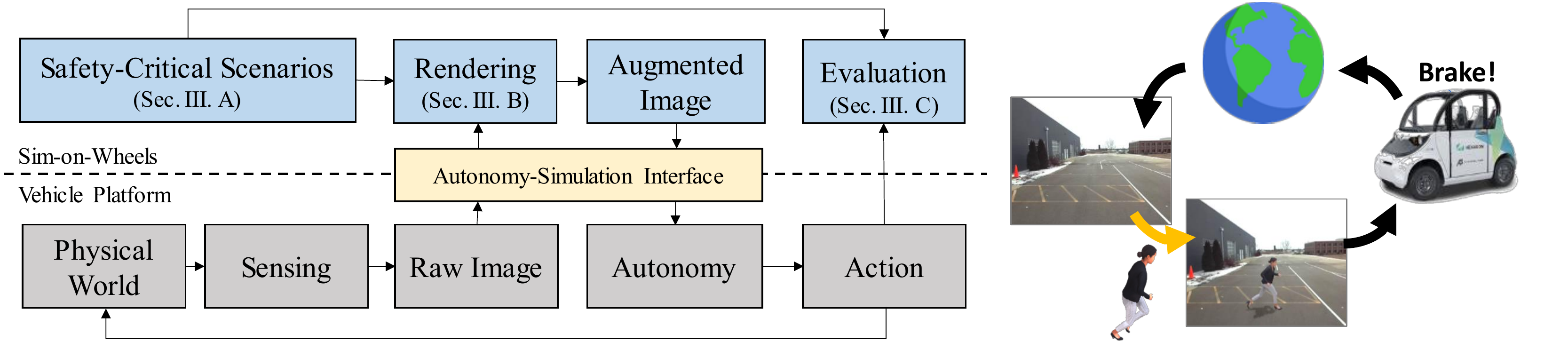}
\caption{
{\bf Sim-on-Wheels Pipeline.} In Sim-on-Wheels' evaluation paradigm, vehicle autonomy is evaluated on images that are perceived in the real world but transmitted to the onboard simulator and manipulated in real-time to show important and dangerous traffic scenarios. The autonomy is asked to react to the manipulated sensory input as if the scenario actually happened. Onboard evaluation can be conducted in real-time to verify the safety and effectiveness of the autonomy. }
\vspace{-4mm}
\label{fig:overview}
\end{figure*}

Evaluating how a self-driving car performs in dangerous scenarios is hard.  Pure real-world
evaluations create situations that are dangerous to participants, while pure simulation evaluations may simulate various scenarios inaccurately, such as cases in which the vehicle has extreme control inputs.  This paper describes
a mixed method, Sim-on-Wheels.  In Sim-on-Wheels, we run actual autonomy stack on real cars, but create scenarios by inserting people and objects into the sensor feed in real-time.  This means we can evaluate the autonomy stack in scenarios
known to be dangerous to pedestrians without risking harm because the pedestrians are simulated.  Furthermore, we
apply the control inputs to a real vehicle.  If the autonomy could cause an uncontrolled skid, we will be able
to measure that. Fig.~\ref{fig:overview} illustrates our evaluation pipeline and Tab.~\ref{tab:related} compares it
to previous methods. In contrast to previous approaches, Sim-on-Wheels is simultaneously safe, interactive, realistic,
and easy to use.

\def\halfcheckmark{\checkmark\kern-1.1ex\raisebox{.7ex}{\rotatebox[origin=c]{125}{--}}}

\begin{table*}
\vspace{1em}
\setlength{\tabcolsep}{4pt}
\centering
\small
\begin{tabular}{lccccc}
\toprule
                                    & \bf Realistic Sensor & \bf Real-world Physics & \bf Closed-loop & \bf Safe & \bf Convenient \\
Off-policy datasets~\cite{geiger2012we, sun2020scalability} 
& \checkmark   &       &          & \checkmark & \checkmark \\
Real-world: road test~\cite{pomerleau1988alvinn, kusano2022collision, cruise-rides}  
& \checkmark        & \checkmark & \checkmark            &   &  \\
Real-world: test track~\cite{ trc, ict}  
& \checkmark        & \checkmark   & \checkmark         & \halfcheckmark &   \\
Simulation: CG-based~\cite{dosovitskiy2017carla, nvdrivesim, shah2018airsim} 
&          &          & \checkmark     &  \checkmark & \checkmark \\
Simulation: data-driven~\cite{manivasagam2020lidarsim, chen2021geosim, tan2021scenegen, amini2022vista, wangcadsim} 
& \halfcheckmark        &         & \checkmark     &  \checkmark & \checkmark \\
Sim-on-Wheels (ours)  
& \checkmark        & \checkmark     & \checkmark        & \checkmark & \checkmark   \\
\bottomrule
\end{tabular}
\caption{A reliable self-driving vehicle evaluation framework requires providing realistic sensors, realistic physics, and closed-loop interaction, all while being safe and easy to use. We situate past frameworks along these five dimensions and discuss them in Section~\ref{sec:related}. 
}
\label{tab:related}
\end{table*}

There is no current consensus on evaluation protocols for autonomous vehicles.
Safety evaluation is typically through a combination of \textbf{real-world
  road-tests},  \textbf{off-policy data collection}, and \textbf{computer simulation}.
Real-world testing is a resource-intensive and risky process, and testing in some
scenarios is unethical (because there is a strong chance of hurting a participant).  Annoyingly, these are the cases
where evaluation is particularly important. Off-policy data can be an effective tool for training and evaluating perception
algorithms, but does not yield a closed-loop evaluation of the safety of the entire autonomy
stack. Computer simulation is safe and scalable, but is not currently reliable in
extreme physical and mechanical situations.  Sim-on-Wheels is a mashup of real-world road tests (so we can observe true vehicle behavior) and computer simulation (so we don't have to risk harm to participants).

An ideal self-driving evaluation environment should be safe, realistic, and closed-loop.
Achieving \textbf{safe} evaluation is challenging, because one should be evaluating dangerous scenarios but
experiments that pose a risk to life are unethical.   \textbf{Realistic} evaluation is essential -- we need to be
sure that evaluation predictions reflect real-life behavior. Finally, \textbf{closed-loop} evaluation is essential
because we must evaluate interactions between the environment and the end-to-end perception to action behavior of the
controller.  Sim-on-Wheels is intrinsically safe and closed-loop.  We show that Sim-on-Wheels results are realistic by both
evaluating the realism of the inserted objects and by comparing conclusions on real and Sim-on-Wheels
scenarios (Section~\ref{sect:reality_gap}). We use Sim-on-Wheels to “spoof” a total of 40 variations of safety-critical scenarios using two different
autonomous vehicle %
pipelines (Section~\ref{exp:autonomy_benchmark}). With the capability of testing scenarios configured at system limit, our Sim-on-Wheel framework reveals our modular agent is more cautious than our end-to-end learned agent in terms of obstacle avoidance, achieving a lower collision rate but taking longer to reach the goals.

\section{Related Work}
\label{sec:related}
\noindent\textbf{Self-driving autonomy:} 
There are two paradigms for self-driving autonomy: (i) modular~\cite{thrun2006stanley,
  doi:10.1126/scirobotics.aat4983} and (ii) end-to-end~\cite{9310544}. 
A modular stack has multiple sub-tasks in a pipeline framework, including localization~\cite{levinson2011towards,
  pmlr-v87-barsan18a}, perception~\cite{chen2017multi, li2022bevformer, liang2018deep},
planning~\cite{kelly2003reactive}, and control~\cite{thrun2006stanley, camacho2013model, johnson2005pid}.
Attractions include interpretability, modularity, and versatility~\cite{9310544}, but tuning the pipeline can
be challenging, and errors can propagate. End-to-end autonomy directly maps sensor input to planner or controller
commands~\cite{muller2005off, zeng2019end, bojarski2016end}. End-to-end methods are usually easier to develop, but tend
to be difficult to interpret, so it is hard to diagnose errors, establish safety
guarantees, and incorporate traffic rules~\cite{bojarski2017explaining}. Recent advances in end-to-end learnable
pipeline autonomy have demonstrated promising results by combining the strengths of both paradigms~\cite{zeng2019end,
  zeng2020dsdnet}. Sim-on-Wheels is designed to be agnostic to the
modular/end-to-end choice.  We evaluate both kinds of autonomy stack in Section~\ref{exp:autonomy_benchmark}.

\noindent\textbf{Self-driving evaluation:}
Evaluation practice is shaped by an important tension between safety and accuracy (accurate
evaluation requires assessment of dangerous scenarios, posing risks to life).
One strategy is to use off-policy datasets~\cite{geiger2012we, sun2020scalability},
which are safe and convenient.  But because
they do not close the perception-action loop, such evaluations cannot accurately
assess a full autonomy stack. Another is to use real-world road tests~\cite{pomerleau1988alvinn, kusano2022collision, cruise-rides}. 
These are expensive
and pose large risks to safety~\cite{ubercrash}, and so are necessarily limited in scope.
Road tests on test tracks~\cite{ trc, ict} 
are somewhat
safer than actual road tests, but are expensive to set up and necessarily provide relatively little environment diversity.  

Yet another is to use a simulator, which is safe and convenient.
Simulated sensor inputs (as in~\cite{dosovitskiy2017carla, nvdrivesim,  shah2018airsim}) 
  face a sim2real gap, despite significant literature on improving the realism of
simulation (e.g. data-driven simulation in~\cite{manivasagam2020lidarsim, chen2021geosim, amini2022vista, wangcadsim};
  dynamic models in~\cite{hofer2021sim2real,
  peng2018sim, andrychowicz2020learning}
; environments in~\cite{tan2021scenegen, suo2021trafficsim,
  feng2022trafficgen, sun2022intersim}). It is extremely difficult to be sure that a simulator captures all relevant
physical modeling. This is particularly important in dangerous scenarios, where one expects extreme control inputs and
odd physics may become important.  For example, reverted rubber hydroplaning is an effect where very aggressive braking
causes tire rubber to break down and capture a surface water film that breaks contact with the road; this and similar
effects can make an important contribution to whether a stack is safe, but may not appear in simulators. Though this could
be dealt with by adding modeling capacity to simulators, it remains difficult to know what to add and when to stop.
In contrast, Sim-on-Wheels uses a real vehicle (and so relies on nature for these effects) but simulates dangerous
scenarios (and so does not endanger participants).

\noindent\textbf{Vehicle-in-the-loop simulation} 
Sim-on-Wheels generally is a vehicle in the loop simulation, because it incorporates the entire vehicle into the
test. Early such methods use a simulated driving environment~\cite{bokc2007validation, albers2010implementation, drechsler2022dynamic, hoenig2015mixed},
with attendant sim2real problems. MiRE~\cite{funk2021mixed} improves realism by using a body tracking system to
map a human into the scene to act as a pedestrian, but the environment is far from realistic.
AR on LiDAR ~\cite{genevois2022augmented} inserts objects in a perceptually realistic
manner into LiDAR point clouds (but not RGB images).  WIL~\cite{hildebrandt2021world}
is a general framework for integrating simulated sensor inputs and real inputs, but does not attend to rendering realism.
In contrast, Sim-on-Wheels provides realistic rendering aimed at specific, safety-critical scenarios.

\begin{table*}[]
    \vspace{1em}
    \centering
    \setlength\tabcolsep{1.0pt}
    \begin{tabular}{ccccc}
    \footnotesize
        \includegraphics[trim={7cm 0 7cm 0},clip, height=0.145\linewidth]{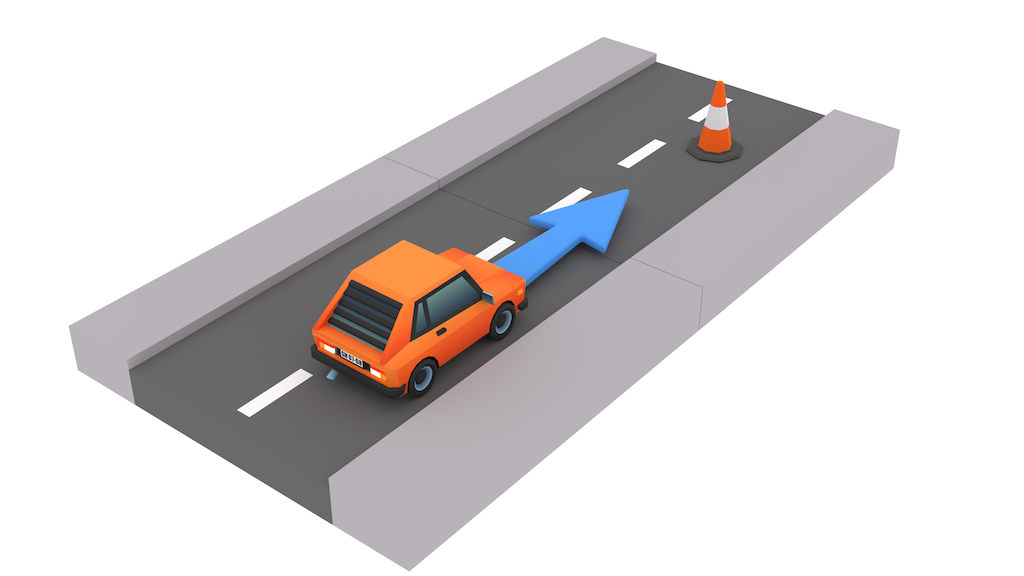} & 
        \includegraphics[trim={7cm 0 7cm 0},clip,height=0.145\linewidth]{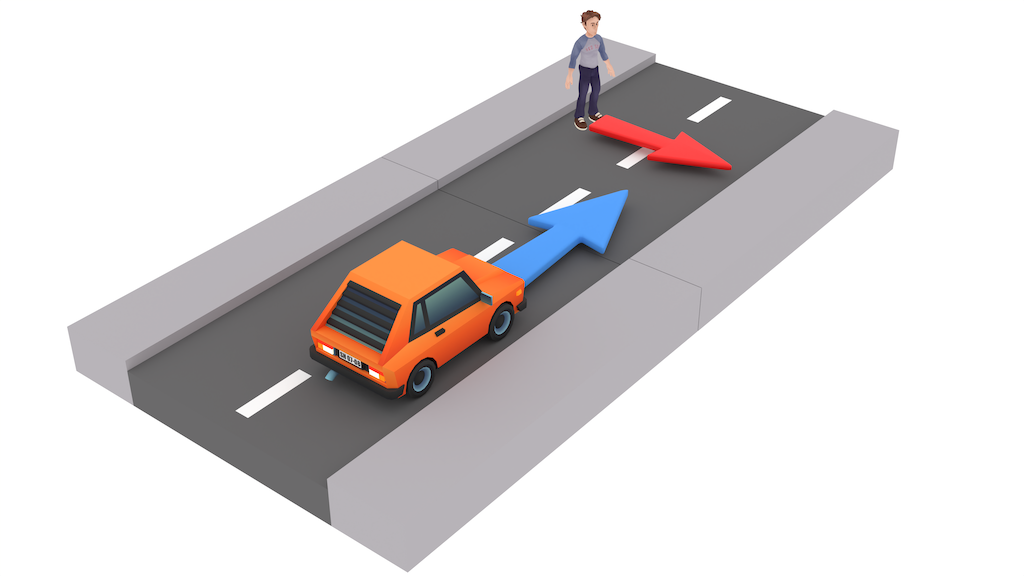} & 
        \includegraphics[trim={7cm 0 7cm 0},clip,height=0.145\linewidth]{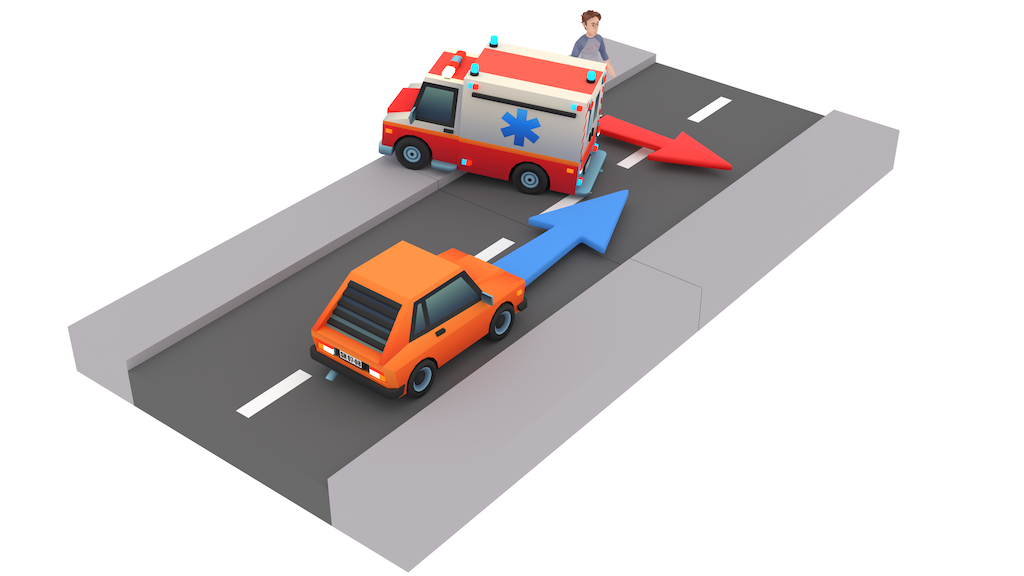} &
        \includegraphics[trim={7cm 0 7cm 0},clip,height=0.145\linewidth]{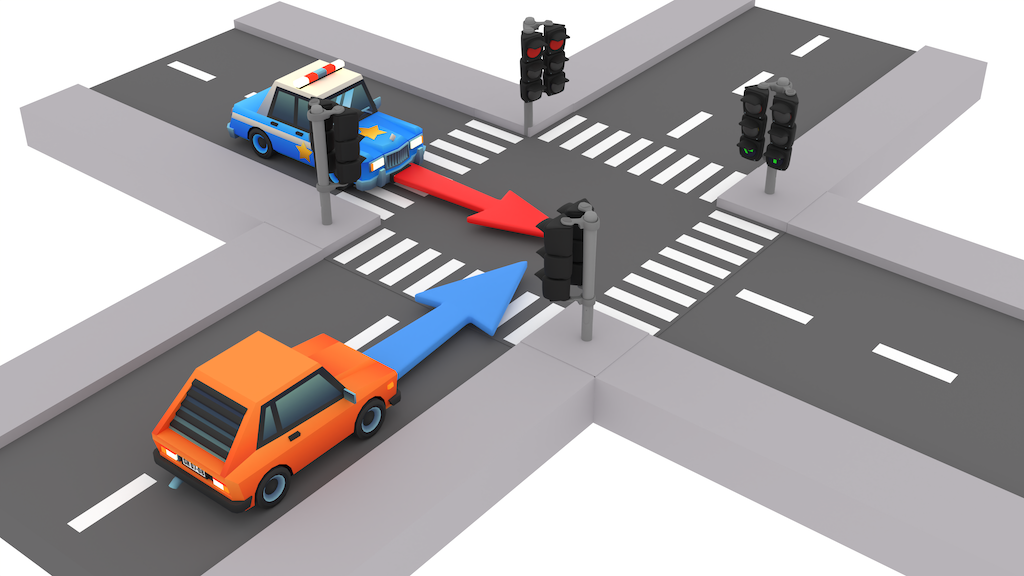} &
        \includegraphics[height=0.145\linewidth]{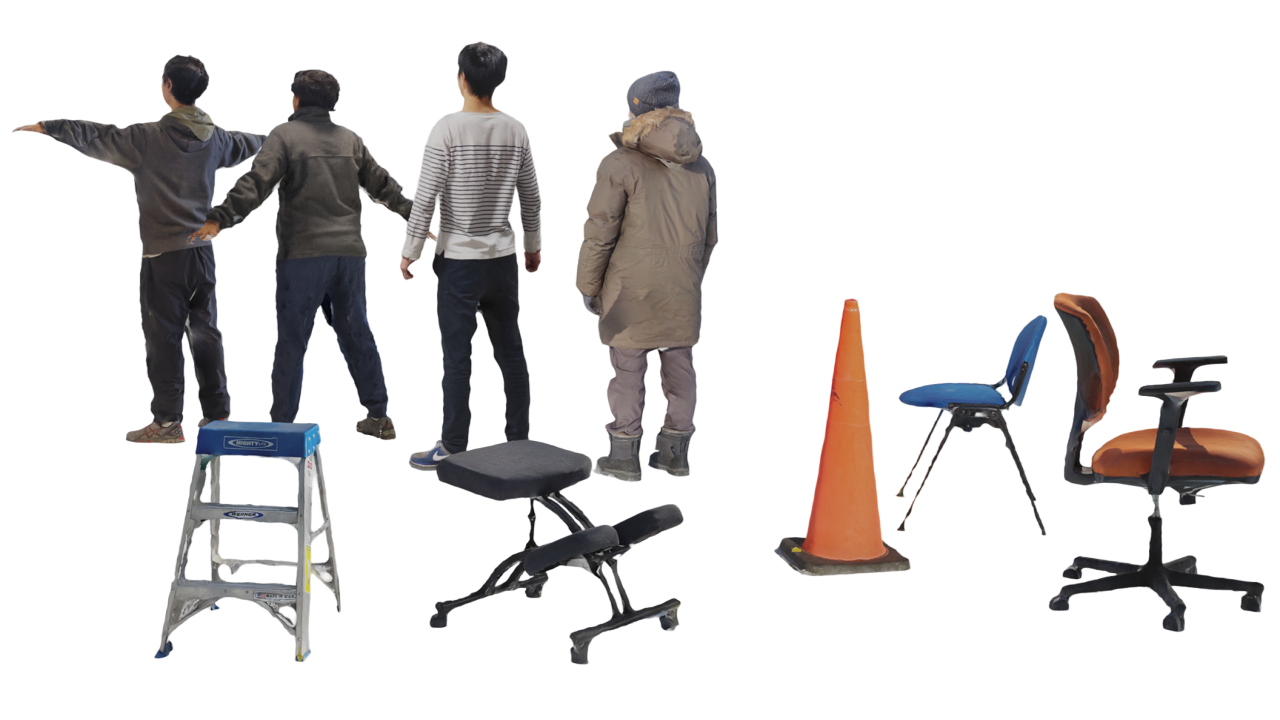}  \\
        Static Obstacle &  
        Jaywalker & 
        Jaywalker w/ Occlusion &
        Traffic light violation & 
        3D assets
    \end{tabular}
    \captionof{figure}{\textbf{Illustration of Testing Scenarios and 3D Assets.} %
    Left: Our scenarios depict common pre-collision events such as road obstacles, jaywalking, jaywalking with occlusions, and traffic light violations. These scenarios are represented as customizable and reproducible spatial-temporal waypoint trajectories for all actors, with triggers, and can be easily expanded upon. Right: We show a subset of our 3D assets, generated from real-world 3D scans using an iPhone equipped with LiDAR.
}
\vspace{-5mm}
    \label{fig:scenarios}
\end{table*}

\section{Simulation in the Physical World}

Sim-on-Wheels operates by inserting actors, objects, and their animations into
the camera stream observed by a controller for a physical autonomous vehicle platform (the
``ego-vehicle'') moving in a real test space. Fig.~\ref{fig:overview} depicts the entire pipeline of our
framework.    There are three main components.
{\bf Authoring:} one must first author a driving scenario to be evaluated (which is likely to be safety critical),
including defining appropriate real-world waypoints and animation sequences for virtual actors/objects, and
determining the planned path for the ego-vehicle to follow (Section~\ref{sec:scenarios}).
{\bf Insertion rendering:} a real-time procedure takes raw RGB-D images,
composites the simulated events into the image stream, and re-publishes the composite images to the agent.
Sufficiently realistic insertion rendering means the ego-vehicle should react in real-time to the inserted objects as if
they were truly present (Section~\ref{sec:rendering}).
{\bf Evaluation:} metrics such as time to goal, stopping distance, and collision rate are computed onboard to evaluate
the effectiveness and safety of different autonomy agents (Section~\ref{sec:evaluation}).

\subsection{Authoring Safety-Critical Scenarios}\label{sec:scenarios}
We choose safety-critical pre-crash scenarios based on the NHTSA pre-crash event report~\cite{najm2007pre}.
Our scenarios encompass common traffic events such
as static obstacles, jaywalking, jaywalking with occlusions, and traffic light violations. They are modeled as
spatial-temporal waypoint trajectories for all actors, allowing for full reproducibility of each scenario. The testing
is conducted to mimic both rural and urban environments, either in a straight road segment or a four-way
intersection. A selection of the scenarios is depicted in Fig.~\ref{fig:scenarios}. 

Authoring involves selecting from a rich collection of 3D assets, including artist-designed assets from SketchFab~\cite{sketchfab}
and in-house created assets reconstructed using an iPhone and multi-view reconstruction software~\cite{realitycapture}. For each scenario, the human actors are animated by Mixamo~\cite{mixamo} with realistic and diverse human animations, such as walking and running. 

The evaluation procedure involves triggering each scenario as the ego-vehicle reaches the trigger zone at a certain
speed range. A crash will occur if the vehicle fails to conduct any evasive action. The hyper-parameters of each
scenario can be adjusted to control the level of difficulty, including the type and trajectory of static objects for the
static obstacle scenarios, the type and trajectory of traffic light runners and trigger distance for the intersection
scenarios, and the walking speed, type and the number of jaywalkers and trigger distance for the jaywalking scenarios,
etc. Our scenario bank can be easily expanded to cover additional safety-critical events. One unique advantage of
Sim-on-Wheels is that it enables setting aggressive hyper-parameters without any risk for physical harm to any vehicles or
pedestrians, and the results of the evaluations provide a comprehensive understanding of the performance limits of each
autonomy stack. Another is that the effect of (say) actor motion or dress on outcomes can be assessed by evaluating the
same scenario for different instances of each actor.

\subsection{Insertion Rendering}\label{sec:rendering}

{\em Insertion rendering} involves producing realistic frames of a scene by inserting assets (image fragments; 3D
models; etc.) into a target image (variants
in~\cite{lalonde2007photoclipart,Liao:2018io,Karsch:sa:11,Karsch14,Liao2012}).
Realistic frames can be produced very fast {\em if}
difficulties presented by lighting, shadows, and geometrical consistency can be managed. Once the scenario is triggered,
we render the simulated scenarios and compose them into images in real time. This requires the insertion rendering to be
realistic, efficient, and geometrically consistent. To achieve this, we adopt a real-time OpenGL-based rasterization pipeline~\cite{Dombi2020}. 

We first place the object accurately in the predefined world coordinate, and the camera pose is acquired in
real-time through an RTK-INS localization module. The lighting consists of the skybox and the sunlight; parameters
are inferred from real-time weather, GPS, and the time of day. 

The rendering process is then conducted using a customized physical-based rendering (PBR) shader that follows the
split-sum shading model, as described by the equation: 
 $
     L(\bx, \bomega_o) = L_a + L_s(\bomega_s)f_r(\bx, \bomega_o,\bomega_s)(\bomega_s \cdot \bn)
 $
, where $x$ is the observed point, $\bomega_o$ is the outgoing ray and $\bomega_s$ is the incoming ray. $L(\bx,
\bomega_o)$ is observed radiance; $L_a$ is ambient sky color and $L_s$ is the directional sunlight; $f_r$ is the Cook-Torrence reflection model~\cite{cook1982reflectance}:  $f_r = k_d f_{d} + k_s f_{s}$, $f_{d}$ is diffuse
reflection under sunlight, and $f_{s}$ describes specular reflection, which accesses the base color, roughness, and
metallic textures of the object to compute specular reflection. The resulting output, as shown in
Fig.~\ref{fig:rendering_quality}, exhibits a visually appealing surface appearance.  

Shadows cast by the inserted objects contribute to the perceived realism. In our framework, a two-pass shadow mapping procedure is applied~\cite{williams1978casting}. The first pass renders a depth buffer from the lighting source to the visible
surface, and the second pass renders per-view depth from the camera perspective. The inconsistent depth between the two
identifies shadowed areas, and Poisson sampling is used to reduce aliasing effects. In addition, occlusion reasoning is
conducted by comparing the rendered depth and perceived depth from the stereo cameras, ensuring the correct depth
ordering between the foreground objects and the background scene. Finally, the rendered objects are composited into the
perceived image through alpha-channel blending.

\begin{table*}[t]
    \vspace{1em}
  \centering
  \footnotesize
  \setlength\tabcolsep{0.05em}
  {
    \begin{tabular}{ccccccc}
    \rotatebox{90}{\phantom{spacingss}Real} &
       \includegraphics[width=.16\linewidth]{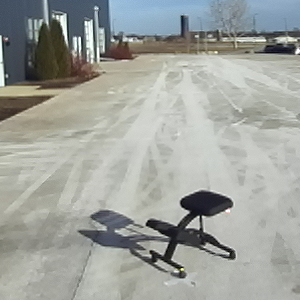} 
       & \includegraphics[width=.16\linewidth]{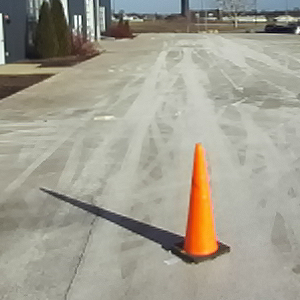} 
       & \includegraphics[width=.16\linewidth]{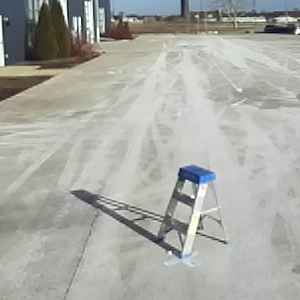}
       & \includegraphics[width=.16\linewidth]{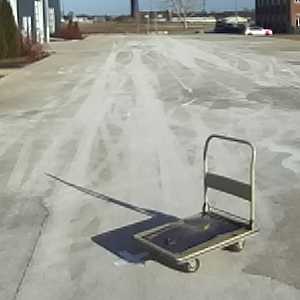}
       & \includegraphics[width=.16\linewidth]{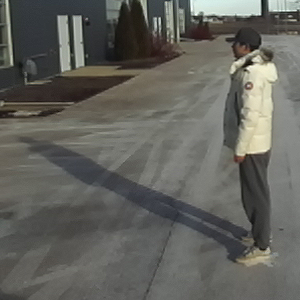}
       & \includegraphics[width=.16\linewidth]{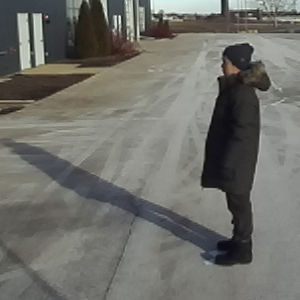} \\
       \rotatebox{90}{\phantom{spacin}Simulation} &
       \includegraphics[width=.16\linewidth]{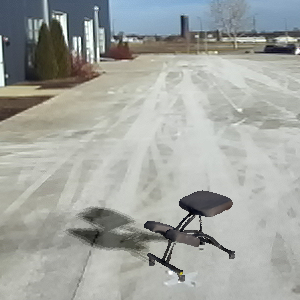} %
        & \includegraphics[width=.16\linewidth]{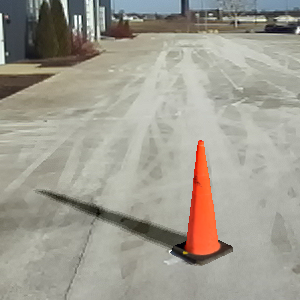}
        & \includegraphics[width=.16\linewidth]{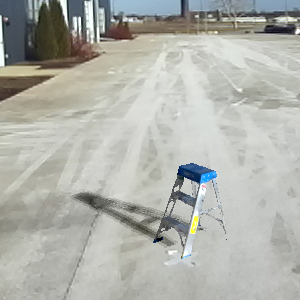} & \includegraphics[width=.16\linewidth]{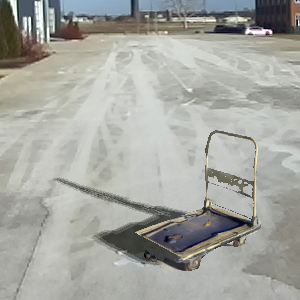}
        & \includegraphics[width=.16\linewidth]{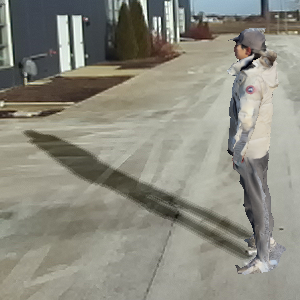} &  \includegraphics[width=.16\linewidth]{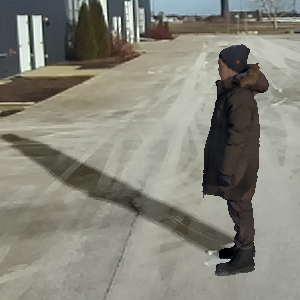}
    \end{tabular}
  }
   \captionof{figure}{\textbf{Rendering Quality.} We evaluate the quality of insertion rendering by comparing reconstructed 3D humans/objects in Sim-on-Wheels with their real-world counterparts under the same pose. The results demonstrate that our real-time insertion rendering can produce realistic, high-fidelity appearances, and cast-shadows. Tab.~\ref{tab:reality-gap} quantitatively measures the sim2real gap. 
   }
   \label{fig:rendering_quality}
   \vspace{-5mm}
\end{table*}

\subsection{Evaluation}\label{sec:evaluation}
The performance of an autonomous driving stack is being evaluated in a simulated safety-critical scenario using recorded
behaviors. We adopt the evaluation metrics from the CARLA platform \cite{dosovitskiy2017carla}, which include the
collision rate and the trip completion time. The collision rate represents the percentage of scenarios where the
ego-vehicle experiences at least one collision, which is determined by checking for overlap between the oriented
bounding box of the vehicle and other virtual objects. In addition to ensuring safety, we aim for our autonomous vehicle
to be as efficient as possible, which is reflected in the goal-reaching time. During each run, the goal-reaching time is measured until the vehicle is
within 5m of a static obstacle, or in other scenarios, within 1.5m of the end of its planned path. In any scenario, if the vehicle is not able to reach its goal, we
penalize that run by recording its time metric as 100s. Furthermore, in order to account for
real-world uncertainties, we report the mean collision rate and goal-reaching time over multiple runs under different
hyper-parameters for one safety-critical scenario.

\begin{table*}[btp!]
    \vspace{1em}
    \centering
    \resizebox{\textwidth}{!}{
    \begin{tabular}{cccc}
    \footnotesize

        Static Obstacle & 
        Jaywalker & 
        Jaywalker with Occlusion & 
        Traffic Light Violation
\\
    \includegraphics[width=.23\linewidth]{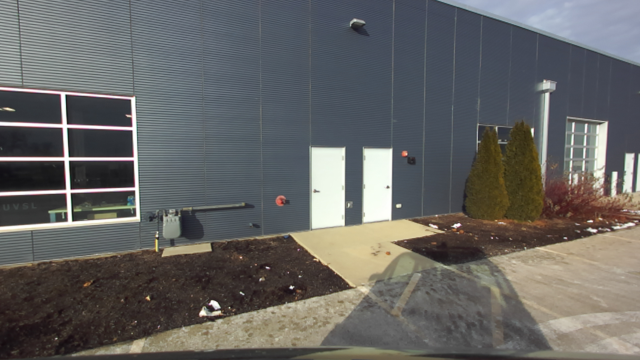}
    & 
    \includegraphics[width=.23\linewidth]{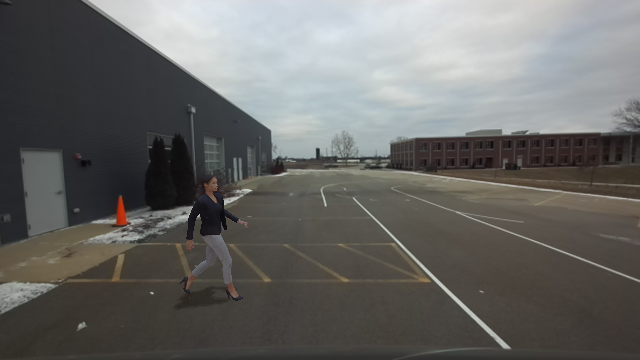}
    & 
    \includegraphics[width=.23\linewidth]{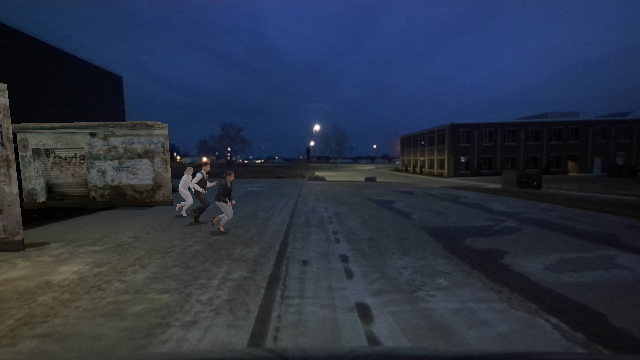}
    &
    \includegraphics[width=.23\linewidth]{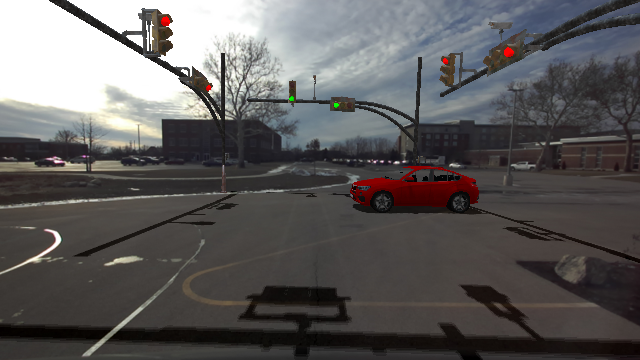} 
    \vspace{-5mm}
    \\
    \includegraphics[height=.25\linewidth, trim={1cm .8cm .8cm .8cm},clip]{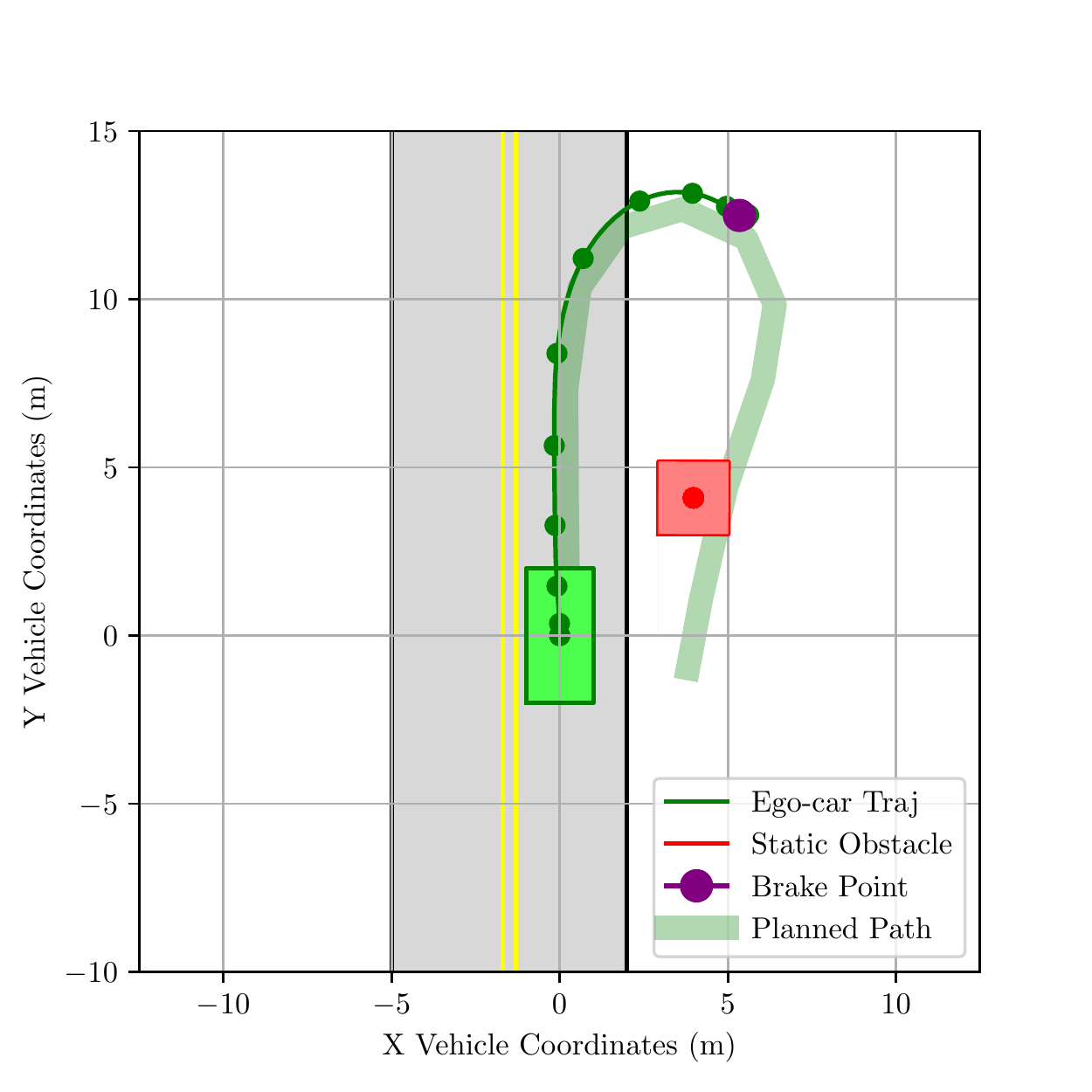}
    & 
        \includegraphics[height=.25\linewidth, trim={1cm .8cm .8cm .8cm},clip]{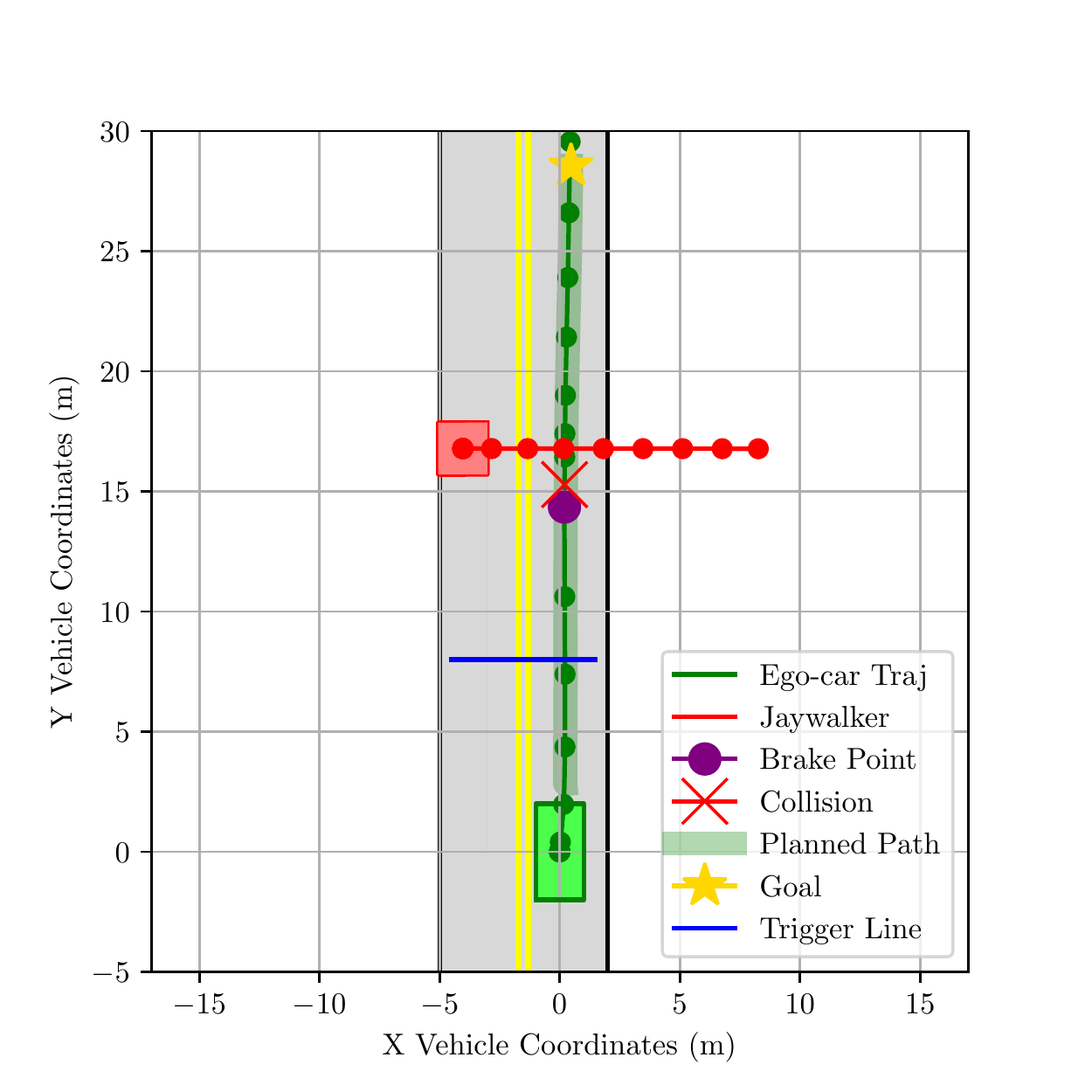}
    & 
    \includegraphics[height=.25\linewidth, trim={1cm .8cm .8cm .8cm},clip]{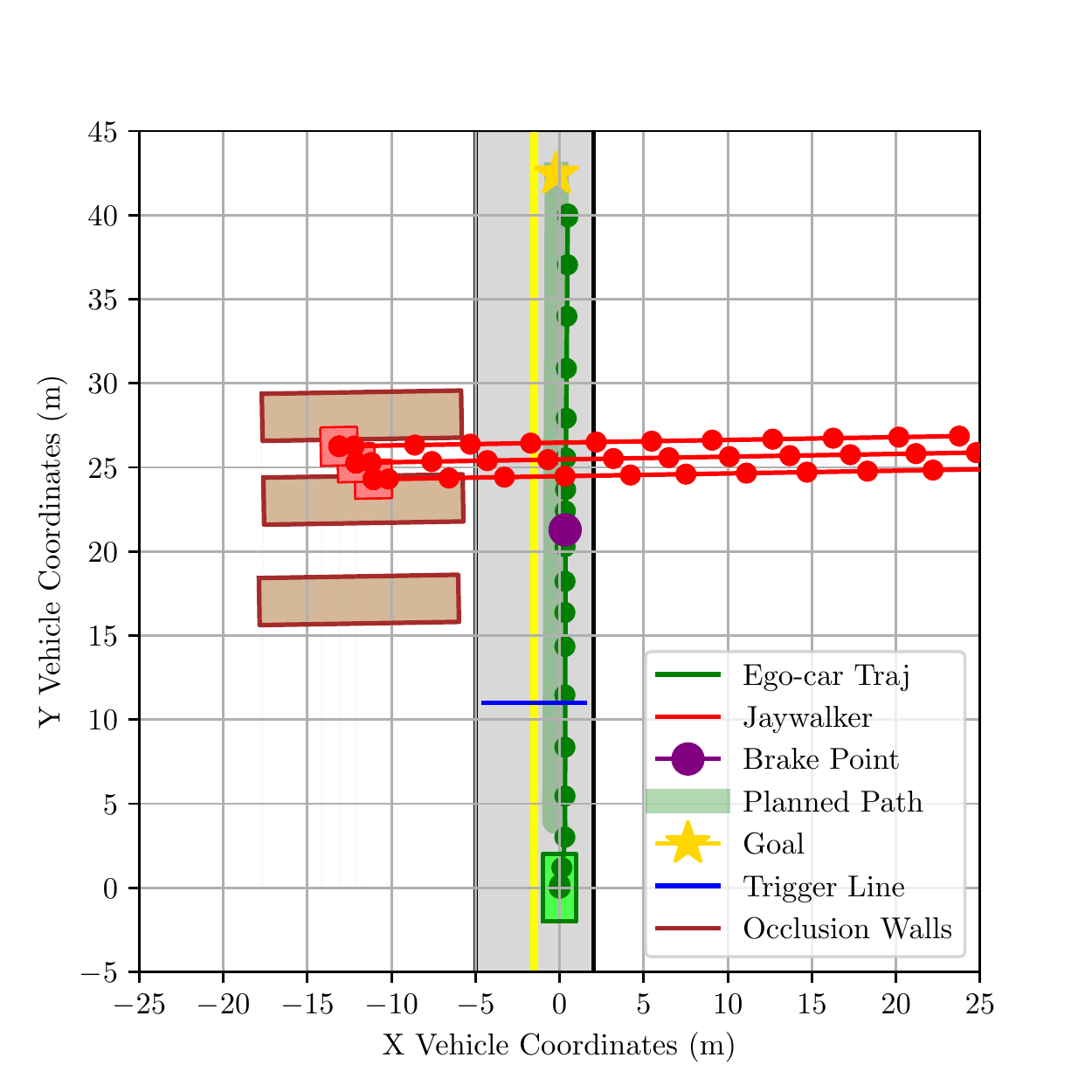}
    & 
    \includegraphics[height=.25\linewidth, trim={1cm .8cm .8cm .8cm},clip]{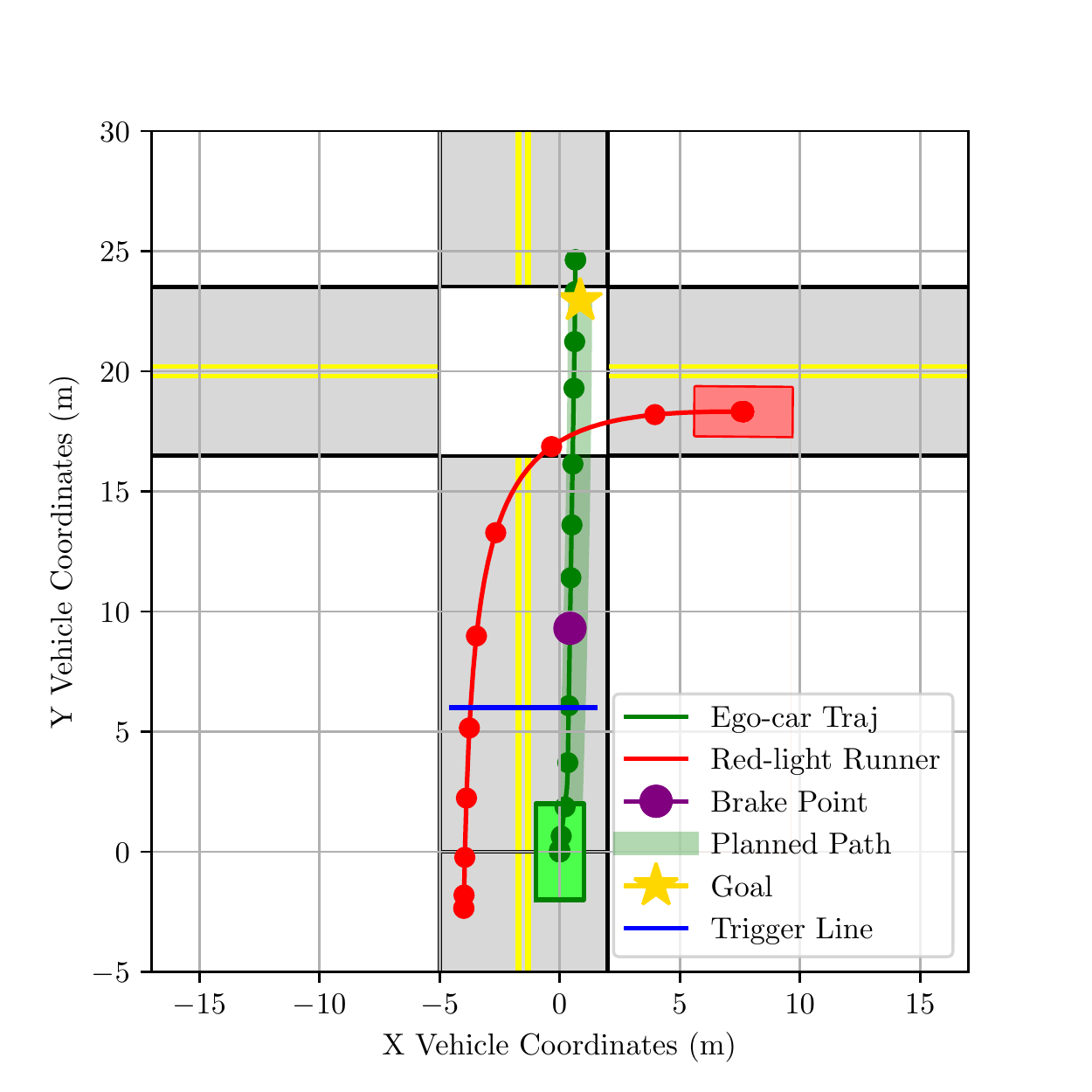}
    \\
    Imitation Learning Agent & 
    Imitation Learning Agent &  
    Imitation Learning Agent &  
    Imitation Learning Agent \\ 
    Reached Goal: {\color{red}No}; Collision: {\color{green}No} 
    & Reached Goal: {\color{green} Yes}; Collision: {\color{red}Yes} 
    & Reached Goal: {\color{green} Yes}; Collision: {\color{green}No} 
    & Reached Goal: {\color{green} Yes}; Collision: {\color{green}No} \vspace{3mm} \\
    
    \includegraphics[width=.23\linewidth]{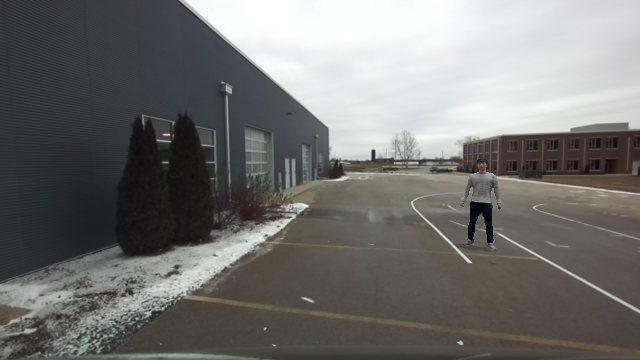}
    & 
    \includegraphics[width=.23\linewidth]{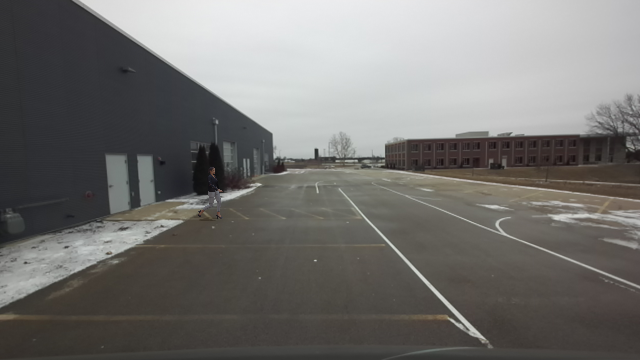}
    & 
    \includegraphics[width=.23\linewidth]{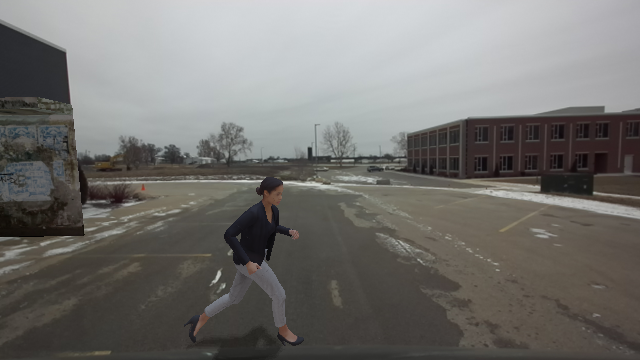}
    &
    \includegraphics[width=.23\linewidth]{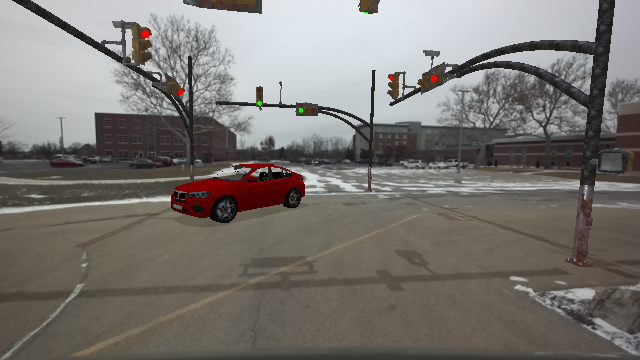} 
    \vspace{-6mm}
    \\
    \includegraphics[height=.25\linewidth, trim={1cm .8cm .8cm .8cm},clip]{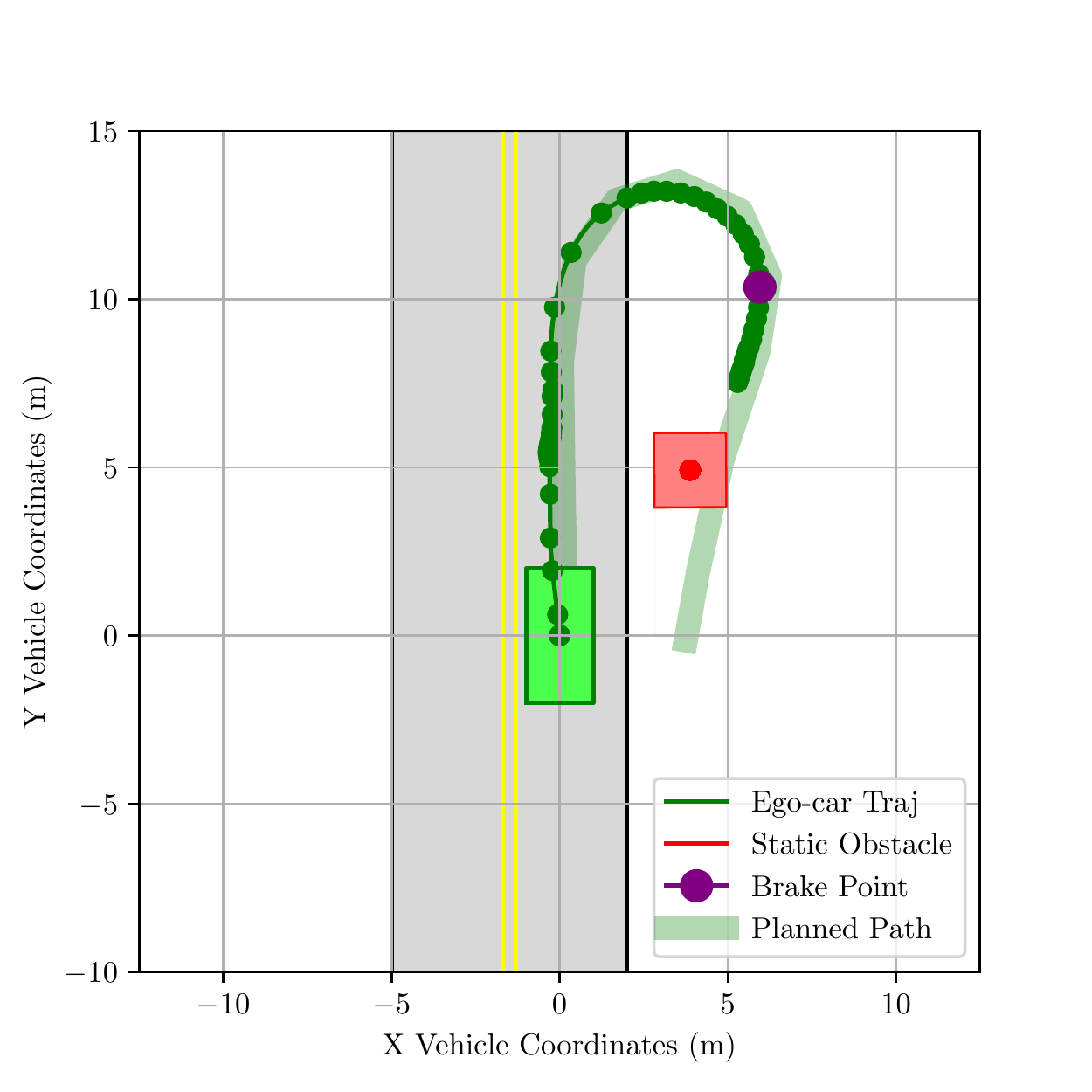}
    & 
        \includegraphics[height=.25\linewidth, trim={1cm .8cm .8cm .8cm},clip]{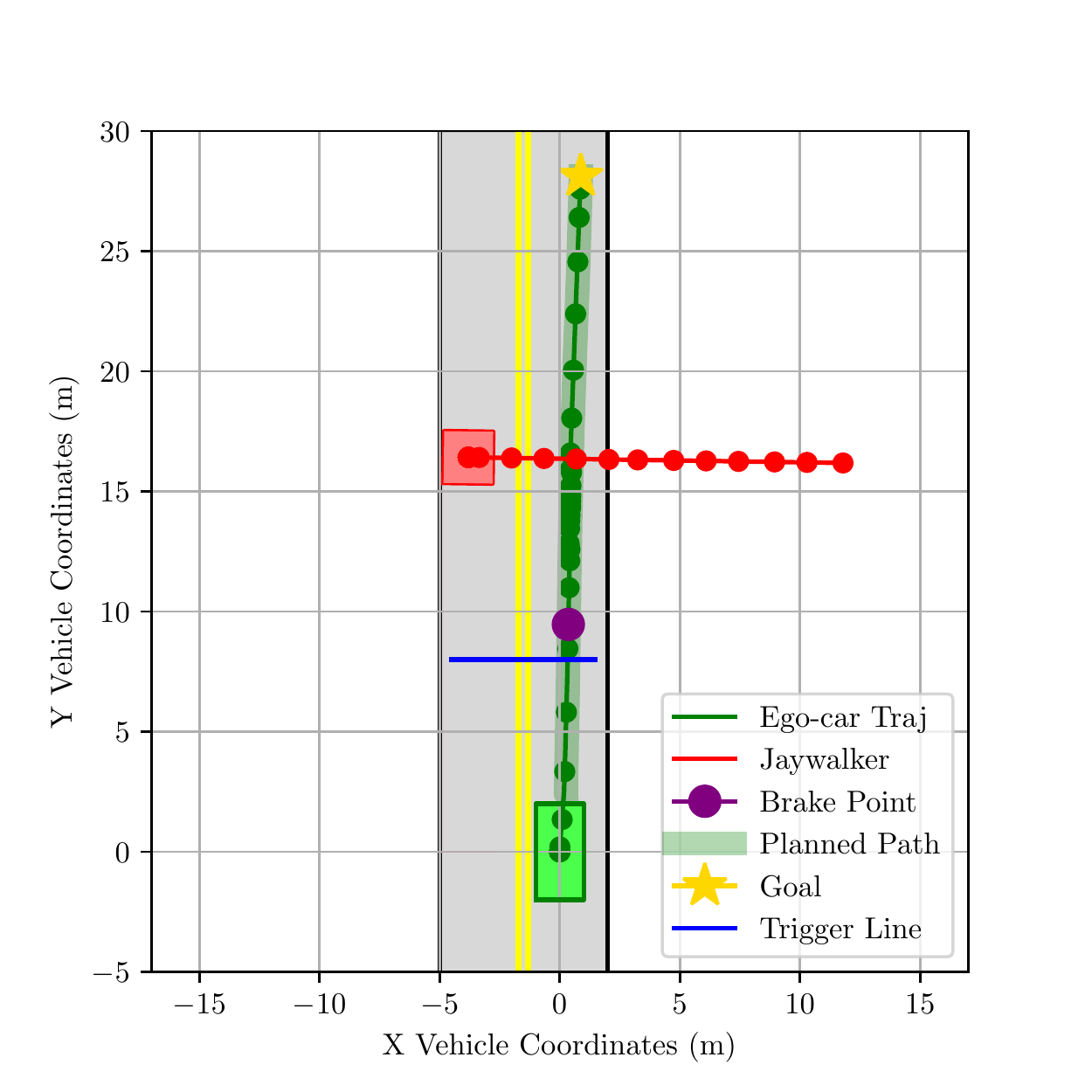}
    & 
    \includegraphics[height=.25\linewidth, trim={1cm .8cm .8cm .8cm},clip]{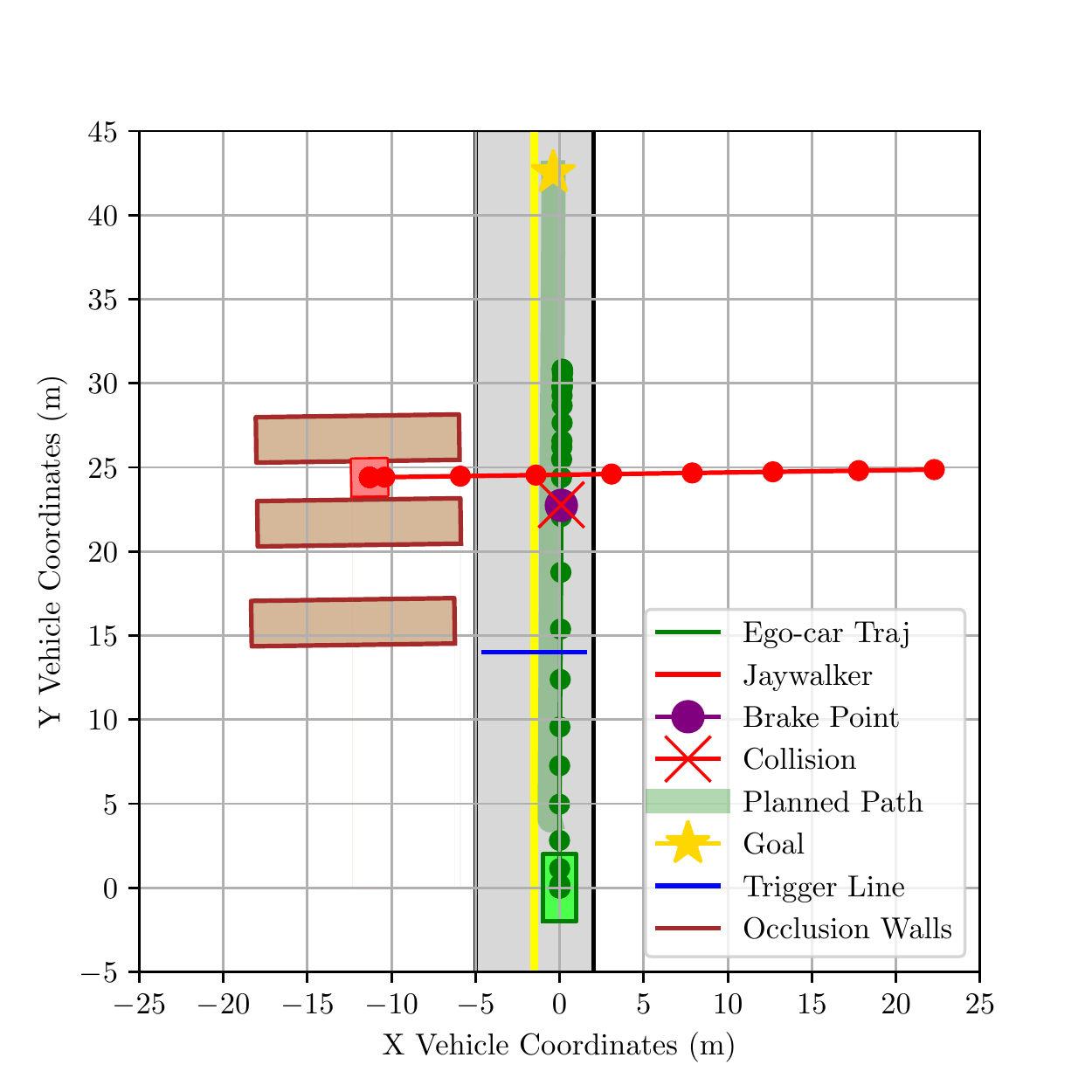}
    & 
    \includegraphics[height=.25\linewidth, trim={1cm .8cm .8cm .8cm},clip]{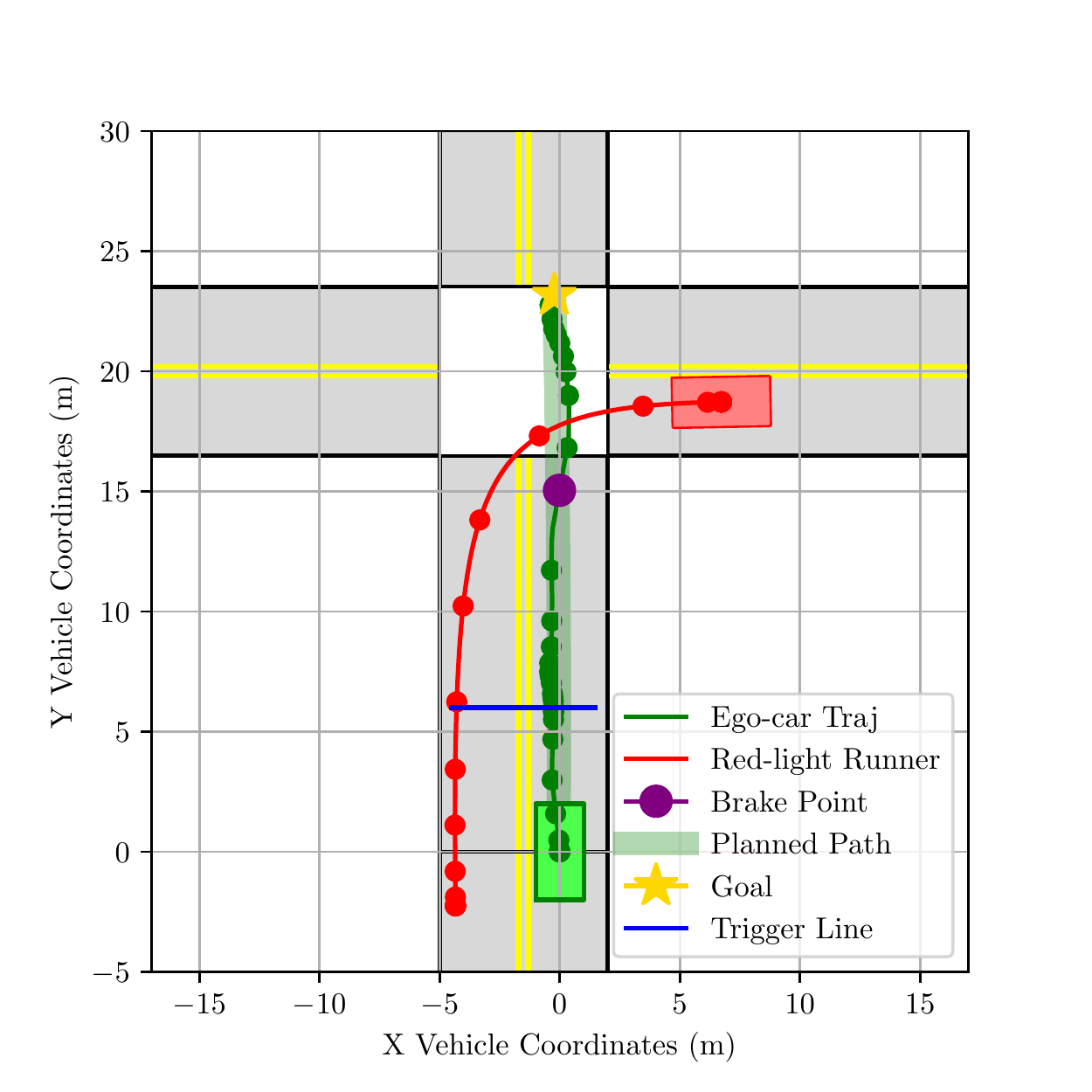}
    \\
    Modular Agent  &  
    Modular Agent  &  
    Modular Agent  &  
    Modular Agent \\
    Reached Goal: {\color{green} Yes}; Collision: {\color{green}No} 
    & Reached Goal: {\color{green} Yes}; Collision: {\color{green}No} 
    & Reached Goal: {\color{red} No}; Collision: {\color{red}Yes}
    & Reached Goal: {\color{green} Yes}; Collision: {\color{green}No}
    \end{tabular}
    }
    \captionof{figure}{\textbf{Qualitative driving results.} : We show bird's-eye view layouts of four different scenarios across both agents in ego-vehicle coordinates (meters). The images %
    are captured at the first braking point before approaching an obstacle. 
    We can see that the modular agent tends to be more cautious and takes longer to reach the goal, while the imitation learner drives smoothly but brakes too early/late in certain scenarios. 
    }
    \label{fig:bev_result}
\end{table*}

\begin{table*}[hbt!]
\centering
\def\arraystretch{1.28}

{
\small
\scalebox{0.74}{
        \hskip-1em\begin{tabular}{c|cc|cc|cc|cc}
        \specialrule{.2em}{.1em}{.1em}
        & \multicolumn{2}{c|}{\textbf{Static Obstacle}}  & \multicolumn{2}{c|}{\textbf{Traffic Light Violation}} & \multicolumn{2}{c|}{\textbf{Jaywalker}} & \multicolumn{2}{c}{\textbf{Jaywalker with Occlusion}}  \\
         Agent Type & Collision Rate $\downarrow$ & Time to Obstacle $\downarrow$ & Collision Rate $\downarrow$ & Time To Goal $\downarrow$ & Collision Rate $\downarrow$ & Time To Goal $\downarrow$ & Collision Rate $\downarrow$ & Time To Goal $\downarrow$ \\ \hline
        Modular Agent & \bf 0.0 & 53.33 & \bf 0.0 & 86.02 & \bf 0.125 & 34.35 & 0.96 & 37.81 \\
        Imitation Learning Agent & 0.33 & \bf 27.78 & 0.33 & \bf 22.90 & 0.875 & \bf 19.58 & \bf 0.58 & \bf 29.41 \\
        \specialrule{.1em}{.05em}{.05em}
        \end{tabular}
        }
}
\caption{{\bf Sim-on-Wheels Benchmark Results}: These metrics are computed as described in Section~\ref{sec:evaluation}. All time metrics above are measured in seconds.
{\bf Findings}: The results show that the modular pipeline is better overall at avoiding collisions, but the imitation learning pipeline reaches the goal faster.  
}
\vspace{-5mm}
\label{tab:benchmark-metrics}
\end{table*}

\section{Experiments}

The goal of the experiments in this section is to address the following three crucial questions: (1) Can the Sim-on-Wheels framework, as proposed, be utilized as a rigorous and comprehensive benchmark for evaluating the performance of various autonomous stacks? (2) Can we empirically validate the authenticity of our simulation?  (3) To what extent does the onboard simulation result in an increase in latency?

In this section, we first provide an overview of the hardware platform and the test track used for our experiments. We then benchmark the performance of two self-driving agents in various safety-critical scenarios using the Sim-on-Wheels framework and conduct a comprehensive analysis of their performance. To quantify the reality gap between the simulation and the real world, we conduct an empirical analysis of the gaps in sensory data and their impact on the perception and action output of the onboard autonomy. Finally, we provide analysis and discussions of our framework. 

\subsection{Real-World Testbed}\label{sec:platform}
All of our experiments are carried out on the Polaris GEM e2, a street-legal, two-seater electric vehicle with a top speed of 25 mph. The sensor stack of the vehicle includes a Velodyne-16 lidar, a Novatel RTK GNSS+INS unit, a ZED 2 Stereo Camera, and a Delphi ESR 2.5 Radar. The vehicle also supports drive-by-wire through the PACMod kit, enabling steering, acceleration, and braking through software.
Our experiments utilize the AStuff Spectra 2~\cite{AStuffSpectra}, an industrial-grade edge computing platform equipped with an Nvidia A4000 GPU. This computer is connected to a built-in monitor located on the passenger side dashboard, facilitating the ease of running and debugging software directly in-vehicle.  

The experiments were carried out in a shared testing track facility, where the testbed area was secured using cones and traffic tape to restrict public access and guarantee the safety of all individuals involved. At all times, a designated safety driver was present behind the wheel of the vehicle, while another team member was stationed outside the vehicle as a lookout, prepared to respond in the event of an emergency. 

\subsection{Evaluated Autonomous Agents}

We subject two distinct autonomous agents for evaluation using the Sim-on-Wheels framework: (1) a modular autonomy stack, and (2) an end-to-end imitation learning stack.

\subsubsection{Modular autonomy} 
Our modular autonomy pipeline takes as input the RGB-D image stream and a coarse planned path. It is composed of four components: detection, tracking, motion prediction, and rule-based longitudinal planning. 

Obstacle detection is split into two modules: static obstacles and dynamic traffic participants. Both rely on pretrained segmentation models without any fine-tuning. Static obstacles are detected by applying a pretrained foreground segmentation model~\cite{xu2022pidnet}. Dynamic obstacles are detected by applying a pretrained instance segmentation model~\cite{cheng2022sparse} to extract instance masks for pedestrians and cars. Each instance mask is converted to a 3D position by taking the median $x$ and $y$ coordinates from the associated depth.  

At the tracking stage, greedy matching~\cite{dyer1993average} is performed to associate the latest detected object and existing tracks based on a bird's eye view. We then estimate the state (velocity and position in bird's eye view) through a linear motion model. Using these velocities and the ego-car's current speed and planned trajectory, we predict the positions of each entity at every \SI{0.2}{\second} step up to \SI{10}{\second} into the future and identify potential collisions. At each future time step, we search for collisions within a fixed collision radius (\SI{3}{\metre}) and travel distance threshold (\SI{5}{\metre}). This threshold can be increased to compensate for latency. If a collision is found, we output a desired speed of zero. Otherwise, we output  \SI{2}{\metre/\second}.

\subsubsection{Imitation learning (IL) agent}%
We also train an end-to-end neural controller agent using behavior cloning. The network takes as input the latest eight frames of speed, position, and RGB image information, which are separated by 0.2 seconds. The controller outputs a continuous brake command to be executed 0.2s in the future, accounting for the latency of the simulation and autonomy pipeline.

The network follows a spatial-temporal recurrent neural network architecture~\cite{xia2020periphery, abduljabbar2021short}. We use an FCN~\cite{long2015fully} backbone pre-trained on instance segmentation on the COCO dataset~\cite{cocodataset} to encode object-level information. The visual features are concatenated with other meta-inputs and fed into a gated recurrent unit (GRU) to incorporate temporal information. The decoder is a multi-layer perceptron outputting a brake value.

To train the network, diverse real-world data is collected from human driving where the driver brakes for static objects and jaywalking pedestrians. The training data contains 195 sequences of static-object braking and approximately 200 sequences of dynamic jaywalking. The network learns to minimize the difference between human driving and its own actions using a $L_1$ regression loss, with the Adam optimizer and a learning rate of 5e-4. Weighted sampling is applied to focus on the relevant samples (e.g. snippets right before brakes), and data augmentation techniques, such as random color jittering and random cropping, are used to increase robustness. We also apply a dropout rate of 0.8 during training.

\subsubsection{Vehicle controller} 
The output from both the modular agent and the IL agent is sent to the same vehicle controller to produce the final vehicle command. The longitudinal direction is controlled by a proportional-integral (PI) speed controller~\cite{zheng1992practical}. Meanwhile, the lateral direction is controlled by the Stanley controller utilizing a bicycle model~\cite{polack2017kinematic}.

\subsubsection{Human driver reference} 
We also benchmark human driver performance within our Sim-on-Wheel system as a reference. The human driver operates the vehicle, navigating it along the planned path and avoiding collisions within simulated scenarios by monitoring real-time augmented video feeds displayed on the GEM vehicle's screen. To prevent the human driver from prematurely reacting to the expectation of an obstacle, we randomized the spawn point of the moving agent, forcing the driver to react on the fly.

\subsection{Autonomy Benchmark Results\label{exp:autonomy_benchmark}}
Tab.~\ref{tab:benchmark-metrics} reports the performance metrics of our driving agents using the Sim-on-Wheels evaluation framework. The qualitative examples of the representative scenarios are depicted in Fig.~\ref{fig:bev_result}. Our analysis, based on the evaluation results and log replays, highlights a few key findings.

Our results indicate that the modular agent generally takes longer to reach the destination due to false positive detections of obstacles, leading to intermittent braking. Nevertheless, it is capable of safely reaching the goal in most cases involving static objects and standard jaywalkers. However, the jaywalker-with-occlusion scenario presents a challenge for the modular agent, as there is limited reaction time between the first appearance of the jaywalker and a potential collision. This results in the failure of the agent to brake in time. Furthermore, the challenging textures of the occluding walls causes the perception module to miss detections.

In contrast, our results indicate that the imitation learning agent does not experience intermittent braking and achieves a shorter time to reach the goal. Furthermore, it performs acceptably well in the intersection scenario with vehicles, which was not part of the training data. However, the agent tends to react late, resulting in higher jaywalker collisions. This may be due to latency differences during training and onboard deployment. It is important to note that such drawbacks were not frequently observed during offline validation, highlighting the importance of real-world vehicle-in-the-loop testing.

Note that none of the agents, including the human driver (Tab. \ref{tab:human_benchmark}), achieve a zero collision rate across all the scenarios, which highlights the difficulty of our designed safety-critical scenarios. It is worth mentioning that many of the test cases, particularly the jaywalking ones, are impractical to test in the real world due to safety concerns and can only be physically evaluated with Sim-on-Wheels.

\begin{table}[t]
\vspace{1em}

    \centering
    \resizebox{\linewidth}{!}{
    \begin{tabular}{cc}
        \centering        
{\includegraphics[height=0.3\linewidth]{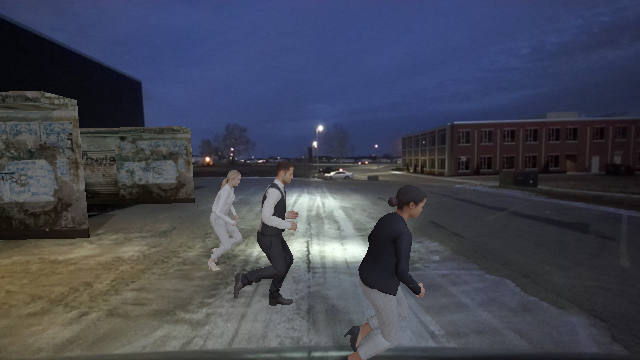}} & \includegraphics[height=.3\linewidth, trim={1cm .8cm .8cm .8cm},clip]{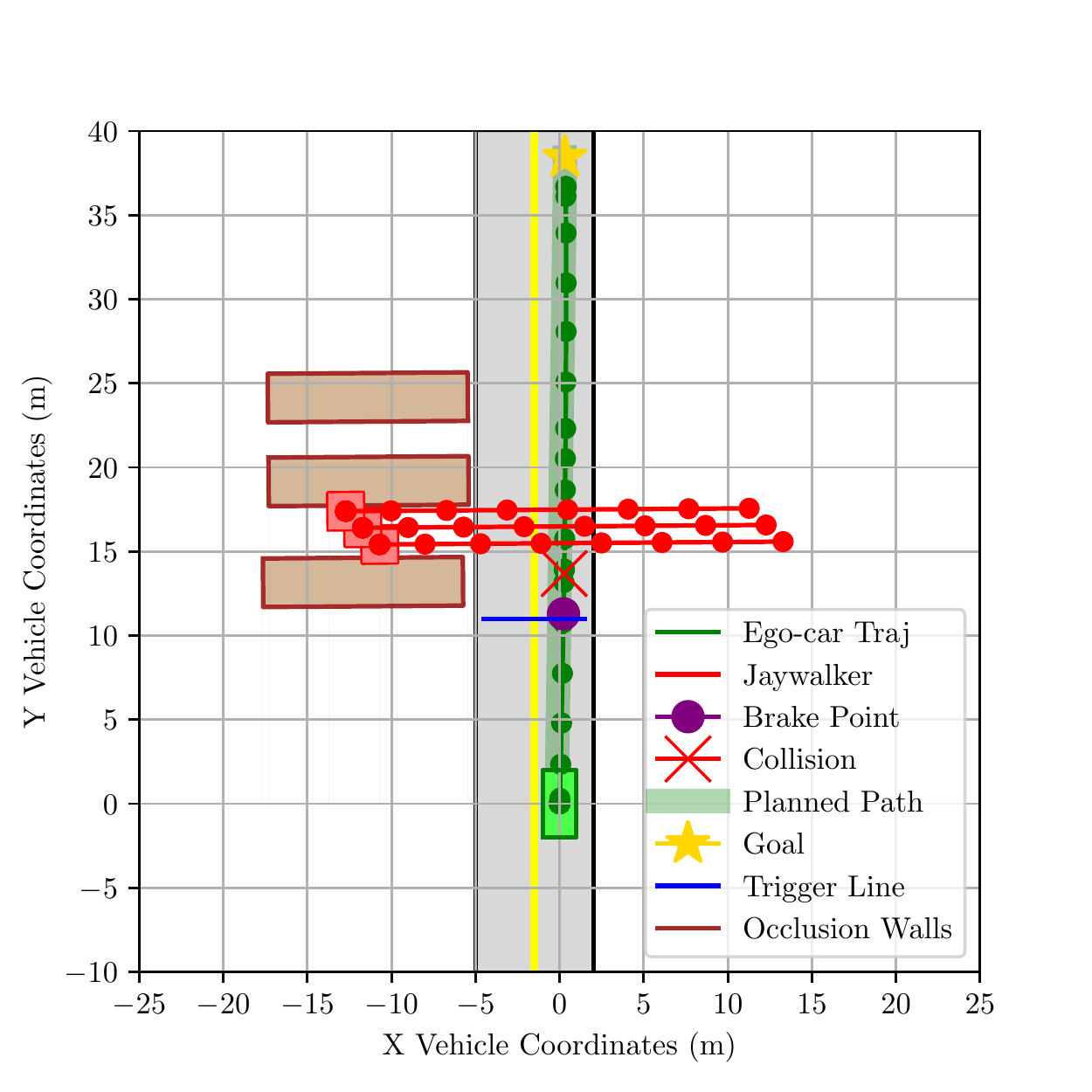}
    \end{tabular}
    }
    \captionof{figure}{\textbf{Human Driving Failure Case}: The one episode in which the human driver collided was the variation of Jaywalker with Occlusion scenario that involved multiple fast jaywalkers. The left image is captured at the braking point.}
    \label{tab:human_failure_vis}
    \vspace{-4mm}
\end{table}
\begin{table}[t]
\centering
\def\arraystretch{1.28}

{
\small
\scalebox{0.8}{
        \begin{tabular}{c|cc}
        \specialrule{.2em}{.1em}{.1em}
         & \multicolumn{2}{c}{\textbf{Jaywalker with Occlusion}}  \\
         Agent Type & Collision Rate $\downarrow$ & Time To Goal $\downarrow$ \\ \hline
        Human & 0.08 & 31.79 \\
        \specialrule{.1em}{.05em}{.05em}
        \end{tabular}
        }
}
\caption{{{\bf Human Driver Benchmark on Sim-on-Wheels}.} These metrics are computed as described in Section~\ref{sec:evaluation}}
\vspace{-5mm}
\label{tab:human_benchmark}
\end{table}

\begin{table*}[hbt!]
\vspace{1em}
\centering
\def\arraystretch{1.28}
{
\small
\scalebox{0.85}{
        \begin{tabular}{cccc|c|cc}
        \specialrule{.2em}{.1em}{.1em}
        \multicolumn{4}{c|}{\textbf{Sensor}} & \multicolumn{1}{c|}{\textbf{Perception}}   & \multicolumn{2}{c}{\textbf{Action}} \\
          PSNR $\uparrow$ & SSIM $\uparrow$ & LPIPS $\downarrow$ & Outlier-$\% \downarrow$ & $\text{mIoU} \uparrow$& $\text{Average Endpoint Offset (m)} \downarrow$ & $\Delta \text{Predicted Brake in } L_1 \downarrow$ \\ \hline
        21.8 & 0.833 & 0.108 & 4.9 & 0.839 & 1.09 $\pm$ 0.49  & 0.18 $\pm$ 0.08 \\
        \specialrule{.1em}{.05em}{.05em}
        \end{tabular}
        }
}
\caption{{\bf Reality Gap}. The reality gap is assessed using three metrics: 1) sensor image fidelity between reality and simulation, 2) mIoU of perception algorithm output, and 3) similarity of final actions (brake values). Results indicate a small reality gap for Sim-on-Wheels, validating its efficacy and reliability as an evaluation framework. }
\label{tab:reality-gap}
\vspace{-4mm}
\end{table*}

\subsection{Reality Gap Analysis}\label{sect:reality_gap}

In Fig~\ref{fig:rendering_quality}, we qualitatively assess our insertion rendering by comparing real vs. simulated results. 
The image pairs appear quite similar overall, demonstrating the robust realism of the framework. However, some minor differences do exist, such as incomplete 3D reconstructed shapes, slight differences in shaded color, and variations in sunlight intensity, shadow shape, and cloud patterns due to the two images being taken in a windy outdoor environment at different times.

Tab.~\ref{tab:reality-gap} reports the quantitative measure using the peak signal-to-noise ratio (PSNR), structural similarity index (SSIM)~\cite{wang2004image}, and a learned perceptual similarity metric (LPIPS)~\cite{zhang2018unreasonable}. The mean absolute error (MAE) of each pixel is calculated and the percentage of outlier pixels with errors over a threshold of $25.5$ under the RGB intensity range $(0, 255)$ were reported. As indicated in Tab.~\ref{tab:reality-gap}, the results show that the virtual object insertions exhibit high fidelity and realism with a low percentage of outlier pixels.

We also evaluate the impact of the reality gap on the performance of our autonomy pipelines. Firstly, we measure the mean Intersection over Union (mIoU) between the static obstacle segmentation network's outputs on real and simulated images. Although the silhouettes of the obstacles match closely, the mIoU is not perfect, which is likely due to the sensitivity of the network to slight variations in color and the background, such as cloud movement. This limitation is a result of the time required to physically arrange obstacles.

Additionally, we evaluate the action-level reality gap by comparing the agent's behavior in real and simulated scenarios. For this, we set the initial vehicle position, planned path, obstacle type, and position to be identical in both scenarios, with only the obstacle being different (real object vs its digital twin). Our experiments show that the modular agent can stop without collision in both real and Sim-on-Wheels experiments. However, we do observe a small discrepancy in its actual trajectory, indicating that the reality gap is not fully closed yet. This may also be due to other factors, such as changes in background illumination as mentioned above.

\subsection{Generalization to other environments}
\begin{table}[t]
  \centering
  \setlength\tabcolsep{0.05em}
  {
    \begin{tabular}{cc}
       \includegraphics[width=.48\linewidth]{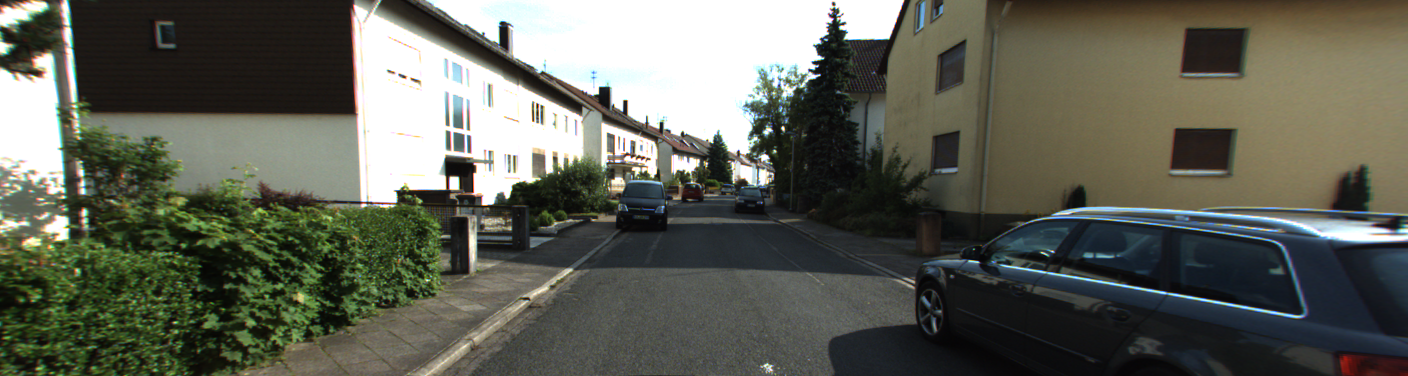} & 
       \includegraphics[width=.48\linewidth]{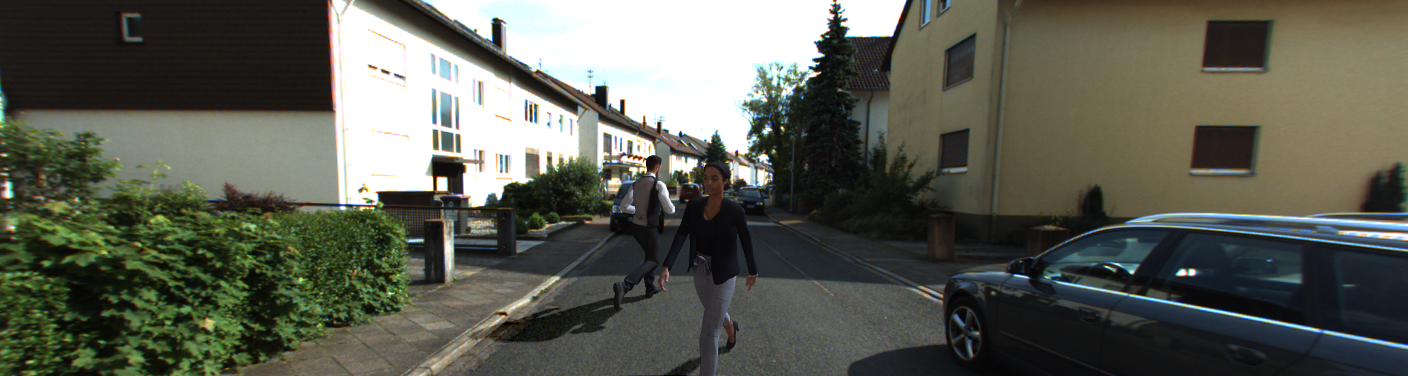} \\
       Original &
       Sim-on-Wheels
    \end{tabular}
  }
   \captionof{figure}{\textbf{Sim-on-Wheels on KITTI-360}: Two jaywalkers on a street are simulated on top of KITTI-360 camera data.}
   \label{fig:kitti}
   \vspace{-4mm}
\end{table}

We can easily adapt to new environments as long as the scenario is compatible with the surroundings. As one example, we generalize to KITTI-360 data~\cite{Liao2022PAMI} as shown Fig. ~\ref{fig:kitti}. 

\begin{table}[hbt!]
\centering
\def\arraystretch{1.28}

{
\small
\scalebox{0.75}{
        \hskip-1em\begin{tabular}{c|ccccc}
        \specialrule{.2em}{.1em}{.1em}
         & Sim-on-Wheels  & Autonomy & ROS & Total &  Relative \%   \\ \hline
         Offline & 43.5ms & - & - & - & - \\
         Online & 63.5ms & - & 212.8ms & 276.3ms & 22.98\\
         Online + IL & 71.5ms & 102.5ms & 214.4ms & 388.4ms & 18.41\\
         Online + Modular & 98.4ms & 129.5ms & 227.3ms & 455.2ms &21.62\\
        \specialrule{.1em}{.05em}{.05em}
        \end{tabular}
}
}
\caption{{{\bf System Runtime Breakdown}. IL stands for imitation learning agent, and modular for modular agent. } %
}
\label{tab:runtime-metrics}
\vspace{-5mm}
\end{table}

\subsection{Runtime Analysis}

The runtime performance of the Sim-on-Wheels system was evaluated on an onboard computer equipped with a single Intel Xeon(R) E-2278G CPU @3.40GHz $\times$16 and a single Nvidia RTX A4000 GPU. The results are presented in Tab.~\ref{tab:runtime-metrics}. The standalone rendering component of the system processes pre-recorded sensor data with a frame rate of 23 FPS. Upon integration with ROS, the frame rate is impacted by the consumption of system resources by hardware drivers, communication modules, the autonomy agent, and the controller. Nonetheless, the system is able to publish rendered images at a minimum frame rate of 10 FPS, even when the autonomy agents run concurrently with the Sim-on-Wheels.

Our results show that the end-to-end latency from the camera capture time to vehicle command increases by 18-21\% with the integration of Sim-on-Wheels, which could be mitigated by equipping the vehicle with two onboard GPUs. Additionally, both autonomy agents can predict future states and actions to compensate for such latency. Tab.~\ref{tab:runtime-metrics} also reveals that the raw ZED2 stereo camera streaming, decoding, and post-processing consumes a substantial amount of time. Despite using separate callback threads to process incoming ROS messages in parallel, a delay still persists due to high-bitrate camera streaming. We plan to further investigate a faster sensor-computer interface to address this issue.

\section{Conclusion}
We propose Sim-on-Wheels, a vehicle-in-the-loop framework for evaluating the performance of autonomous vehicles in real-world scenarios in a safe and realistic manner. To the best of our knowledge, Sim-on-Wheels is the first framework of its kind to support the integration of simulation and real-world testing practices for the safe and realistic evaluation of autonomous vehicles. Our results demonstrate the versatility and reliability of Sim-on-Wheels as a framework for evaluating various agents. To further support the research and development of autonomous driving, we will open-source Sim-on-Wheels to the community and establish a safe, closed-loop, end-to-end, real-world benchmark.
\section*{Acknowledgments}
This project has been funded by the Amazon Research Award, Nvidia Hardware Grants, the Insper-Illinois Innovative Grant, and the NCSA Faculty Fellow. We are grateful for the support of the Center for Autonomy and John M. Hart. We thank Wei-Chiu Ma and Jason Ren for proofreading. 
\bibliographystyle{IEEEtran}
\bibliography{references, extrabib}

\begin{thebibliography}{10}
\providecommand{\url}[1]{#1}
\csname url@samestyle\endcsname
\providecommand{\newblock}{\relax}
\providecommand{\bibinfo}[2]{#2}
\providecommand{\BIBentrySTDinterwordspacing}{\spaceskip=0pt\relax}
\providecommand{\BIBentryALTinterwordstretchfactor}{4}
\providecommand{\BIBentryALTinterwordspacing}{\spaceskip=\fontdimen2\font plus
\BIBentryALTinterwordstretchfactor\fontdimen3\font minus
  \fontdimen4\font\relax}
\providecommand{\BIBforeignlanguage}[2]{{%
\expandafter\ifx\csname l@#1\endcsname\relax
\typeout{** WARNING: IEEEtran.bst: No hyphenation pattern has been}%
\typeout{** loaded for the language `#1'. Using the pattern for}%
\typeout{** the default language instead.}%
\else
\language=\csname l@#1\endcsname
\fi
#2}}
\providecommand{\BIBdecl}{\relax}
\BIBdecl

\bibitem{geiger2012we}
A.~Geiger, P.~Lenz, and R.~Urtasun, ``Are we ready for autonomous driving? the
  kitti vision benchmark suite,'' in \emph{CVPR}, 2012.

\bibitem{sun2020scalability}
P.~Sun, H.~Kretzschmar, X.~Dotiwalla, A.~Chouard, V.~Patnaik, P.~Tsui, J.~Guo,
  Y.~Zhou, Y.~Chai, B.~Caine \emph{et~al.}, ``Scalability in perception for
  autonomous driving: Waymo open dataset,'' in \emph{CVPR}, 2020.

\bibitem{pomerleau1988alvinn}
D.~A. Pomerleau, ``Alvinn: An autonomous land vehicle in a neural network,''
  \emph{NeurIPS}, 1988.

\bibitem{kusano2022collision}
K.~D. Kusano, K.~Beatty, S.~Schnelle, F.~Favaro, C.~Crary, and T.~Victor,
  ``Collision avoidance testing of the waymo automated driving system,''
  \emph{arXiv preprint arXiv:2212.08148}, 2022.

\bibitem{cruise-rides}
``Cruise {D}riverless {R}ides,'' \url{https://getcruise.com/rides}. Accessed at
  2023-02-04.

\bibitem{trc}
``{TRC} - {T}ransportation {R}esearch {C}enter,'' \url{https://www.trcpg.com}.
  Accessed at 2023-02-04.

\bibitem{ict}
``Illinois {A}utonomous and {C}onnected {T}rack,''
  \url{https://ict.illinois.edu/i-act/testing-arena}. Accessed at 2023-02-04.

\bibitem{dosovitskiy2017carla}
A.~Dosovitskiy, G.~Ros, F.~Codevilla, A.~Lopez, and V.~Koltun, ``Carla: An open
  urban driving simulator,'' in \emph{CoRL}, 2017.

\bibitem{nvdrivesim}
``{NVIDIA DRIVE} {S}im,''
  \url{https://developer.nvidia.com/drive/drive-constellation}. Accessed at
  2023-02-04.

\bibitem{shah2018airsim}
S.~Shah, D.~Dey, C.~Lovett, and A.~Kapoor, ``Airsim: High-fidelity visual and
  physical simulation for autonomous vehicles,'' in \emph{Field and Service
  Robotics: Results of the 11th International Conference}, 2018.

\bibitem{manivasagam2020lidarsim}
S.~Manivasagam, S.~Wang, K.~Wong, W.~Zeng, M.~Sazanovich, S.~Tan, B.~Yang,
  W.-C. Ma, and R.~Urtasun, ``Lidarsim: Realistic lidar simulation by
  leveraging the real world,'' in \emph{CVPR}, 2020.

\bibitem{chen2021geosim}
Y.~Chen, F.~Rong, S.~Duggal, S.~Wang, X.~Yan, S.~Manivasagam, S.~Xue, E.~Yumer,
  and R.~Urtasun, ``Geosim: Realistic video simulation via geometry-aware
  composition for self-driving,'' in \emph{CVPR}, 2021.

\bibitem{tan2021scenegen}
S.~Tan, K.~Wong, S.~Wang, S.~Manivasagam, M.~Ren, and R.~Urtasun, ``Scenegen:
  Learning to generate realistic traffic scenes,'' in \emph{CVPR}, 2021.

\bibitem{amini2022vista}
A.~Amini, T.-H. Wang, I.~Gilitschenski, W.~Schwarting, Z.~Liu, S.~Han,
  S.~Karaman, and D.~Rus, ``Vista 2.0: An open, data-driven simulator for
  multimodal sensing and policy learning for autonomous vehicles,'' in
  \emph{ICRA}, 2022.

\bibitem{wangcadsim}
J.~Wang, S.~Manivasagam, Y.~Chen, Z.~Yang, I.~A. B{\^a}rsan, A.~J. Yang, W.-C.
  Ma, and R.~Urtasun, ``Cadsim: Robust and scalable in-the-wild 3d
  reconstruction for controllable sensor simulation,'' in \emph{CoRL}, 2022.

\bibitem{thrun2006stanley}
S.~Thrun, M.~Montemerlo, H.~Dahlkamp, D.~Stavens, A.~Aron, J.~Diebel, P.~Fong,
  J.~Gale, M.~Halpenny, G.~Hoffmann \emph{et~al.}, ``Stanley: The robot that
  won the darpa grand challenge,'' \emph{Journal of field Robotics}, 2006.

\bibitem{doi:10.1126/scirobotics.aat4983}
J.~Daudelin, G.~Jing, T.~Tosun, M.~Yim, H.~Kress-Gazit, and M.~Campbell, ``An
  integrated system for perception-driven autonomy with modular robots,''
  \emph{Science Robotics}, 2018.

\bibitem{9310544}
A.~Tampuu, T.~Matiisen, M.~Semikin, D.~Fishman, and N.~Muhammad, ``A survey of
  end-to-end driving: Architectures and training methods,'' \emph{IEEE
  Transactions on Neural Networks and Learning Systems}, 2022.

\bibitem{levinson2011towards}
J.~Levinson, J.~Askeland, J.~Becker, J.~Dolson, D.~Held, S.~Kammel, J.~Z.
  Kolter, D.~Langer, O.~Pink, V.~Pratt \emph{et~al.}, ``Towards fully
  autonomous driving: Systems and algorithms,'' in \emph{IV}, 2011.

\bibitem{pmlr-v87-barsan18a}
I.~A. Barsan, S.~Wang, A.~Pokrovsky, and R.~Urtasun, ``Learning to localize
  using a lidar intensity map,'' in \emph{CoRL}, 2018.

\bibitem{chen2017multi}
X.~Chen, H.~Ma, J.~Wan, B.~Li, and T.~Xia, ``Multi-view 3d object detection
  network for autonomous driving,'' in \emph{CVPR}, 2017.

\bibitem{li2022bevformer}
Z.~Li, W.~Wang, H.~Li, E.~Xie, C.~Sima, T.~Lu, Y.~Qiao, and J.~Dai,
  ``Bevformer: Learning bird’s-eye-view representation from multi-camera
  images via spatiotemporal transformers,'' in \emph{ECCV}, 2022.

\bibitem{liang2018deep}
M.~Liang, B.~Yang, S.~Wang, and R.~Urtasun, ``Deep continuous fusion for
  multi-sensor 3d object detection,'' in \emph{ECCV}, 2018.

\bibitem{kelly2003reactive}
A.~Kelly and B.~Nagy, ``Reactive nonholonomic trajectory generation via
  parametric optimal control,'' \emph{IJRR}, 2003.

\bibitem{camacho2013model}
E.~F. Camacho and C.~B. Alba, \emph{Model predictive control}.\hskip 1em plus
  0.5em minus 0.4em\relax Springer science \& business media, 2013.

\bibitem{johnson2005pid}
M.~A. Johnson and M.~H. Moradi, \emph{PID control}.\hskip 1em plus 0.5em minus
  0.4em\relax Springer, 2005.

\bibitem{muller2005off}
U.~Muller, J.~Ben, E.~Cosatto, B.~Flepp, and Y.~Cun, ``Off-road obstacle
  avoidance through end-to-end learning,'' \emph{NeurIPS}, 2005.

\bibitem{zeng2019end}
W.~Zeng, W.~Luo, S.~Suo, A.~Sadat, B.~Yang, S.~Casas, and R.~Urtasun,
  ``End-to-end interpretable neural motion planner,'' in \emph{CVPR}, 2019.

\bibitem{bojarski2016end}
M.~Bojarski, D.~Del~Testa, D.~Dworakowski, B.~Firner, B.~Flepp, P.~Goyal, L.~D.
  Jackel, M.~Monfort, U.~Muller, J.~Zhang \emph{et~al.}, ``End to end learning
  for self-driving cars,'' \emph{arXiv preprint arXiv:1604.07316}, 2016.

\bibitem{bojarski2017explaining}
M.~Bojarski, P.~Yeres, A.~Choromanska, K.~Choromanski, B.~Firner, L.~Jackel,
  and U.~Muller, ``Explaining how a deep neural network trained with end-to-end
  learning steers a car,'' \emph{arXiv preprint arXiv:1704.07911}, 2017.

\bibitem{zeng2020dsdnet}
W.~Zeng, S.~Wang, R.~Liao, Y.~Chen, B.~Yang, and R.~Urtasun, ``Dsdnet: Deep
  structured self-driving network,'' in \emph{ECCV}, 2020.

\bibitem{ubercrash}
N.~T.~S. Board, ``Collision between vehicle controlled by developmental
  automated driving system and pedestrian,'' \emph{NTSB Highway Accident
  Report}, 2019.

\bibitem{hofer2021sim2real}
S.~H{\"o}fer, K.~Bekris, A.~Handa, J.~C. Gamboa, M.~Mozifian, F.~Golemo,
  C.~Atkeson, D.~Fox, K.~Goldberg, J.~Leonard \emph{et~al.}, ``Sim2real in
  robotics and automation: Applications and challenges,'' \emph{IEEE
  transactions on automation science and engineering}, 2021.

\bibitem{peng2018sim}
X.~B. Peng, M.~Andrychowicz, W.~Zaremba, and P.~Abbeel, ``Sim-to-real transfer
  of robotic control with dynamics randomization,'' in \emph{ICRA}, 2018.

\bibitem{andrychowicz2020learning}
O.~M. Andrychowicz, B.~Baker, M.~Chociej, R.~Jozefowicz, B.~McGrew,
  J.~Pachocki, A.~Petron, M.~Plappert, G.~Powell, A.~Ray \emph{et~al.},
  ``Learning dexterous in-hand manipulation,'' \emph{IJRR}, 2020.

\bibitem{suo2021trafficsim}
S.~Suo, S.~Regalado, S.~Casas, and R.~Urtasun, ``Trafficsim: Learning to
  simulate realistic multi-agent behaviors,'' in \emph{CVPR}, 2021.

\bibitem{feng2022trafficgen}
L.~Feng, Q.~Li, Z.~Peng, S.~Tan, and B.~Zhou, ``Trafficgen: Learning to
  generate diverse and realistic traffic scenarios,'' \emph{arXiv preprint
  arXiv:2210.06609}, 2022.

\bibitem{sun2022intersim}
Q.~Sun, X.~Huang, B.~C. Williams, and H.~Zhao, ``Intersim: Interactive traffic
  simulation via explicit relation modeling,'' in \emph{IROS}, 2022.

\bibitem{bokc2007validation}
T.~Bokc, M.~Maurer, and G.~Farber, ``Validation of the vehicle in the loop
  (vil); a milestone for the simulation of driver assistance systems,'' in
  \emph{IV}, 2007.

\bibitem{albers2010implementation}
A.~Albers and T.~D{\"u}ser, ``Implementation of a vehicle-in-the-loop
  development and validation platform,'' in \emph{FISITA World automotive
  congress}, 2010.

\bibitem{drechsler2022dynamic}
M.~F. Drechsler, V.~Sharma, F.~Reway, C.~Sch{\"u}tz, and W.~Huber, ``Dynamic
  vehicle-in-the-loop: A novel method for testing automated driving
  functions,'' \emph{SAE International Journal of Connected and Automated
  Vehicles}, 2022.

\bibitem{hoenig2015mixed}
W.~Hoenig, C.~Milanes, L.~Scaria, T.~Phan, M.~Bolas, and N.~Ayanian, ``Mixed
  reality for robotics,'' in \emph{IROS}, 2015.

\bibitem{funk2021mixed}
M.~Funk~Drechsler, J.~B. Peintner, G.~Seifert, W.~Huber, and A.~Riener, ``Mixed
  reality environment for testing automated vehicle and pedestrian
  interaction,'' in \emph{13th International Conference on Automotive User
  Interfaces and Interactive Vehicular Applications}, 2021.

\bibitem{genevois2022augmented}
T.~Genevois, J.-B. Horel, A.~Renzaglia, and C.~Laugier, ``Augmented reality on
  lidar data: Going beyond vehicle-in-the-loop for automotive software
  validation,'' in \emph{IV}, 2022.

\bibitem{hildebrandt2021world}
C.~Hildebrandt and S.~Elbaum, ``World-in-the-loop simulation for autonomous
  systems validation,'' in \emph{ICRA}, 2021.

\bibitem{najm2007pre}
W.~G. Najm, J.~D. Smith, M.~Yanagisawa \emph{et~al.}, ``Pre-crash scenario
  typology for crash avoidance research,'' United States. National Highway
  Traffic Safety Administration, Tech. Rep., 2007.

\bibitem{sketchfab}
``Sketchfab search,'' \url{https://sketchfab.com/search?q=bus&amp;type=models}.
  Accessed at 2023-02-04.

\bibitem{realitycapture}
``Realityscan,'' \url{https://www.capturingreality.com/}. Accessed at
  2023-02-04.

\bibitem{mixamo}
{Adobe}, ``Mixamo,'' \url{https://www.mixamo.com}. Accessed at 2023-02-04.

\bibitem{lalonde2007photoclipart}
J.-F. Lalonde, D.~Hoiem, A.~A. Efros, C.~Rother, J.~Winn, and A.~Criminisi,
  ``Photo clip art,'' \emph{TOG}, 2007.

\bibitem{Liao:2018io}
Z.~Liao, K.~Karsch, H.~Zhang, and D.~Forsyth, ``{An Approximate Shading Model
  with Detail Decomposition for Object Relighting},'' \emph{IJCV}, 2018.

\bibitem{Karsch:sa:11}
K.~Karsch, V.~Hedau, D.~Forsyth, and D.~Hoiem, ``{Rendering synthetic objects
  into legacy photographs},'' in \emph{SIGGRAPH Asia}, 2011.

\bibitem{Karsch14}
K.~Karsch, K.~Sunkavalli, S.~Hadap, N.~Carr, H.~Jin, R.~Fonte, M.~Sittig, and
  D.~Forsyth, ``Automatic scene inference for 3d object compositing,''
  \emph{ACM Trans. Graph.}, 2014.

\bibitem{Liao2012}
Z.~Liao, A.~Farhadi, Y.~Wang, I.~Endres, and D.~Forsyth, ``Building a
  dictionary of image fragments,'' in \emph{CVPR}, 2012.

\bibitem{Dombi2020}
S.~Dombi. Moderngl, high performance python bindings for opengl 3.3+.
  \url{https://github.com/moderngl/moderngl}. Accessed at 2023-02-04.

\bibitem{cook1982reflectance}
R.~L. Cook and K.~E. Torrance, ``A reflectance model for computer graphics,''
  \emph{ToG}, 1982.

\bibitem{williams1978casting}
L.~Williams, ``Casting curved shadows on curved surfaces,'' in
  \emph{Proceedings of the 5th annual conference on Computer graphics and
  interactive techniques}, 1978.

\bibitem{AStuffSpectra}
``Astuff spectra 2,''
  \url{https://autonomoustuff.com/products/astuff-spectra-2}. Accessed at
  2023-02-04.

\bibitem{xu2022pidnet}
J.~Xu, Z.~Xiong, and S.~P. Bhattacharyya, ``Pidnet: A real-time semantic
  segmentation network inspired from pid controller,'' \emph{arXiv preprint
  arXiv:2206.02066}, 2022.

\bibitem{cheng2022sparse}
T.~Cheng, X.~Wang, S.~Chen, W.~Zhang, Q.~Zhang, C.~Huang, Z.~Zhang, and W.~Liu,
  ``Sparse instance activation for real-time instance segmentation,'' in
  \emph{CVPR}, 2022.

\bibitem{dyer1993average}
M.~Dyer, A.~Frieze, and B.~Pittel, ``The average performance of the greedy
  matching algorithm,'' \emph{The Annals of Applied Probability}, 1993.

\bibitem{xia2020periphery}
Y.~Xia, J.~Kim, J.~Canny, K.~Zipser, T.~Canas-Bajo, and D.~Whitney,
  ``Periphery-fovea multi-resolution driving model guided by human attention,''
  in \emph{WACV}, 2020.

\bibitem{abduljabbar2021short}
R.~L. Abduljabbar, H.~Dia, P.-W. Tsai, and S.~Liyanage, ``Short-term traffic
  forecasting: An lstm network for spatial-temporal speed prediction,''
  \emph{Future Transportation}, 2021.

\bibitem{long2015fully}
J.~Long, E.~Shelhamer, and T.~Darrell, ``Fully convolutional networks for
  semantic segmentation,'' in \emph{CVPR}, 2015.

\bibitem{cocodataset}
T.~Lin, M.~Maire, S.~J. Belongie, L.~D. Bourdev, R.~B. Girshick, J.~Hays,
  P.~Perona, D.~Ramanan, P.~Doll{'{a} }r, and C.~L. Zitnick, ``Microsoft
  {COCO:} common objects in context,'' \emph{CoRR}, 2014.

\bibitem{zheng1992practical}
L.~Zheng, ``A practical guide to tune of proportional and integral (pi) like
  fuzzy controllers,'' in \emph{[1992 Proceedings] IEEE International
  Conference on Fuzzy Systems}, 1992.

\bibitem{polack2017kinematic}
P.~Polack, F.~Altch{\'e}, B.~d'Andr{\'e}a Novel, and A.~de~La~Fortelle, ``The
  kinematic bicycle model: A consistent model for planning feasible
  trajectories for autonomous vehicles?'' in \emph{IV}, 2017.

\bibitem{wang2004image}
Z.~Wang, A.~C. Bovik, H.~R. Sheikh, and E.~P. Simoncelli, ``Image quality
  assessment: from error visibility to structural similarity,'' \emph{TIP},
  2004.

\bibitem{zhang2018unreasonable}
R.~Zhang, P.~Isola, A.~A. Efros, E.~Shechtman, and O.~Wang, ``The unreasonable
  effectiveness of deep features as a perceptual metric,'' in \emph{CVPR},
  2018.

\bibitem{Liao2022PAMI}
Y.~Liao, J.~Xie, and A.~Geiger, ``{KITTI}-360: A novel dataset and benchmarks
  for urban scene understanding in 2d and 3d,'' \emph{Pattern Analysis and
  Machine Intelligence (PAMI)}, 2022.

\end{thebibliography}
\end{document}